\documentclass[final]{cvpr}
\usepackage{times}
\usepackage{epsfig}
\usepackage{graphicx}
\usepackage{amsmath}
\usepackage{amssymb}
\usepackage{subfigure}
\usepackage{caption}
\usepackage{multirow}
\usepackage{booktabs}
\usepackage{listings}

\usepackage[pagebackref=true,breaklinks=true,colorlinks,bookmarks=false]{hyperref}

\def\cvprPaperID{3226}
\def\confYear{CVPR 2021}

\author{Xinzhu Ma\textsuperscript{1},\ \  
	Yinmin Zhang\textsuperscript{3}, \ \  
	Dan Xu\textsuperscript{2},  \ \
	Dongzhan Zhou\textsuperscript{1}, \\
	Shuai Yi\textsuperscript{3}, \ \
	Haojie Li\textsuperscript{4}, \ \
	Wanli Ouyang\textsuperscript{1} \\  
\textsuperscript{1}The University of Sydney, \ \
\textsuperscript{2}The Hong Kong University of Science and Technology,\\
\textsuperscript{3}SenseTime Research, \ \
\textsuperscript{4}Dalian University of Technology\\

{\tt\small \{xinzhu.ma, d.zhou, wanli.ouyang\}@sydney.edu.au,}
\\
{\tt\small \{zhangyinmin, yishuai\}@sensetime.com,} \ \
{\tt\small danxu@cse.ust.hk,} \ \ 
{\tt\small hjli@dlut.edu.cn}
}

\begin{document}

%%%%%%%%% TITLE
\title{Delving into Localization Errors for Monocular 3D Object Detection}

\maketitle
\pagestyle{empty}  
\thispagestyle{empty}

%%%%   abstract  
%%%%% abstract

\begin{abstract}
Estimating 3D bounding boxes from monocular images is an essential component in autonomous driving, while accurate 3D object detection from this kind of data is very challenging.
In this work, by intensive diagnosis experiments, we quantify the impact introduced by each sub-task and found the `localization error' is the vital factor in restricting monocular 3D detection.
Besides, we also investigate the underlying reasons behind localization errors, analyze the issues they might bring, and propose three strategies.
First, we revisit the misalignment between the center of the 2D bounding box and the projected center of the 3D object, which is a vital factor leading to low localization accuracy.
Second, we observe that accurately localizing distant objects with existing technologies is almost impossible, while those samples will mislead the learned network.
To this end, we propose to remove such samples from the training set for improving the overall performance of the detector.
Lastly, we also propose a novel 3D IoU oriented loss for the size estimation of the object, which is not affected by `localization error'. 
We conduct extensive experiments on the KITTI dataset, where the proposed method achieves real-time detection and outperforms previous methods by a large margin.
The code will be made available at: \url{https://github.com/xinzhuma/monodle}.

\end{abstract}

%%%%   introduction
%%%%  introduction
\vspace{-5pt}
\section{Introduction}
Remarkable progress has been achieved in 3D detection, especially for LiDAR/stereo-based approaches~\cite{zhou2018voxelnet,lang2019pointpillars,shi2019pointrcnn,chen2020dsgn,wang2019pseudo}, along with the advances in deep neural networks.
In contrast, the accuracy of 3D detection from only monocular images~\cite{Simonelli_2019_ICCV,brazil2019m3d,chen2020monopair,Ma_2019_ICCV,Ma_2020_ECCV,Ding_2020_CVPR} is obviously lower than that from LiDAR or stereo. 
In this work, we aim to quantitatively identify the problem and propose our solutions.
%, which is generally attributed to the lack of depth information. 

%Unfortunately, estimating depth from a single RGB image is an ill-posed problem, making monocular 3D object detection an extremely challenging task. From our observations, depth estimation is indeed one of the bottlenecks of this task, but there are still many other factors restricting the performance of monocular 3D detection. In this work, we aim to explore those problems and propose our solutions.

\begin{figure}[t]
\centering
%\fbox{\rule{0pt}{1.7in} \rule{0.\linewidth}{0pt}}
\includegraphics[width=0.99\linewidth]{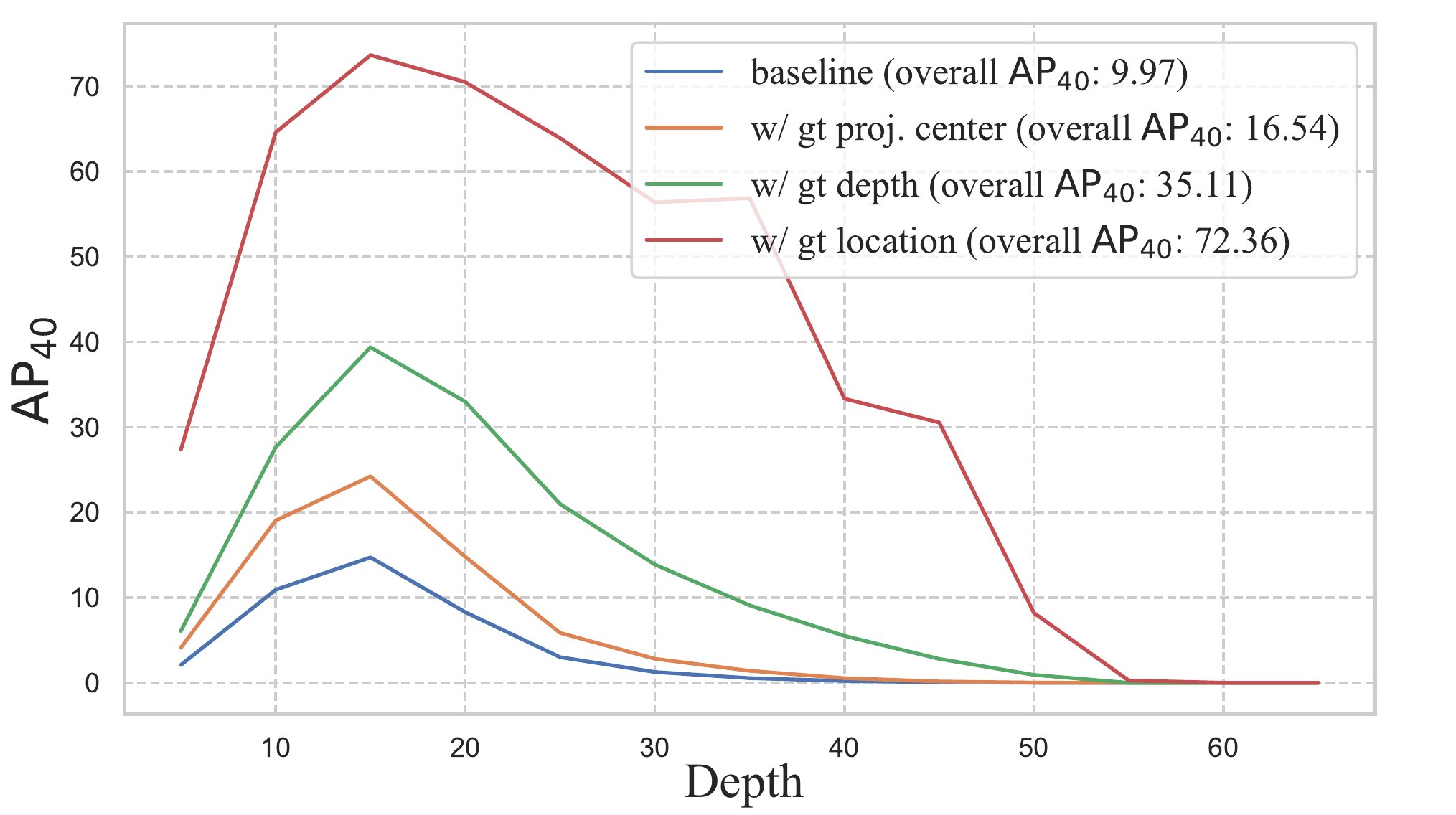}
\vspace{-5pt}
   \caption{{\bf Range-wise evaluation on the KITTI \emph{validation} set.} Metric is ${\rm AP}_{40}$ of the Car category under moderate setting.
   The sampling interval is 10 m. For example, the corresponding value at horizontal axis 20 represents the overall performance of all samples between 15 m and 25 m.}
\label{fig:evalbyrange}
\vspace{-10pt}
\end{figure}

To investigate and quantify the underlying factors that restrict the performance of monocular 3D object detection,
we conduct intensive diagnostic experiments for this task, inspired by the error identifying methods~\cite{Law_2018_ECCV,zhou2019objects,hoiem2012diagnosing,bolya2020tide} commonly used in the 2D detection scope.
Specifically, we build our baseline model (see Section \ref{sec:baseline} for details) based on CenterNet~\cite{zhou2019objects} and progressively replace predicted items with their ground-truth values.
To better analyze the error patterns, we evaluate the results in a range-wise manner and show the summary of those experiments in Figure~\ref{fig:evalbyrange}.
Based on our investigation, we have the following three observations and corresponding designs.

\emph{Observation 1}:
The most striking feature in Figure~\ref{fig:evalbyrange} is the leap in performance when using ground-truth location, reaching a level similar to the state-of-the-art LiDAR-based methods, suggesting the localization error is the key factor in restricting monocular 3D detection.
Furthermore, except for depth estimation, detecting the projected center of the 3D object also plays an important role in restoring the 3D position of the object.
To this end, we revisit the misalignment between the center of the 2D bounding box and the projected center of the 3D object. 
Besides, we also confirm the necessity of keeping 2D detection related branches in monocular 3D detector.
In this way, 2D detection is used as the correlated auxiliary task to help learning the features shared with 3D detection, which is different from the existing work in~\cite{Liu_2020_CVPR_Workshops} that discards 2D detection.

\emph{Observation 2}:
An apparent trend reflected in Figure~\ref{fig:evalbyrange} is that the detection accuracy significantly decreases with respect to the distance (the low performance of very close range objects will be discussed in supplementary materials).
More importantly, all the models cannot output any true positive samples beyond a certain distance.
We found that it is almost impossible to detect distant objects accurately with existing technologies due to the inevitable localization errors (see Section~\ref{sec:experment_analysis} for details).
In this case, whether it is beneficial to add these samples into the training set becomes a question.
In fact, there is a clear domain gap between `bad' samples and `easy-to-detect' samples and forcing the network to learn from those samples will reduce its representative ability for the others, which will thus impair the overall performance. 
Based on the observation above, we propose two schemes. The first scheme removes distant samples from the training set and the second scheme reduces the training loss weights of these samples.

\emph{Observation 3}:
We found that, except for localization error, there are also some other vital factors, such as dimension estimation, restricting monocular 3D detection (there is still 27.4\% room for improvements even we use the ground-truth location).
%\emph{Analysis and design:}
Existing methods in this scope tend to optimize each component of the 3D bounding box independently, and the studies in \cite{Simonelli_2019_ICCV,monodis_tpami}  confirm the effectiveness of this strategy.
However, the failure to consider the contribution of each loss item to the final metric (\ie 3D IoU) may lead to sub-optimal optimization.
To alleviate this problem, we propose an IoU oriented loss for 3D size estimation. The new IoU oriented loss dynamically adjust the loss weight for each side in sample level according its contribution rate to the 3D IoU. 

In summary, the key contributions of this paper are as follows:
% The contributions of this paper can be summarized as:
First, we conduct intensive diagnostic experiments for monocular 3D detection.
In addition to finding that the `localization error' is the main problem restricting monocular 3D detection, we also quantify the overall impact of each sub-task.
Second, we investigate the underlying reasons behind localization error, analyze the issues it might bring.
Accordingly, we propose three novel strategies operating on annotations, training samples, and optimization losses to alleviate problems caused by localization error for boosting the detection.

Experimental results show the effectiveness of the proposed strategies.
In particular, compared with existing best-performing monocular 3D object detection approaches, the proposed method achieves at least 1.6 points ${\rm AP}_{40}$ improvements on the bird's view detection and 3D object detection in the KITTI dataset.
%where the proposed method outperforms the previous state-of-the-art by a large margin and achieves real-time detection. %Coupled with the proposed baseline model and the three improvements mentioned before, we achieve at least 1.6 points ${\rm AP}_{40}$ improvements on nine metrics of the 3D object detection in the KITTI benchmark.

%%%%   components
\section{Related Work}

\noindent
{\bf Standard monocular 3D detection.}
Here we briefly review the ‘standard’ monocular 3D detection approaches \cite{mousavian20173d,roddick2018orthographic,qin2019monogrnet,Simonelli_2019_ICCV,brazil2019m3d,zhou2019objects} only use the RGB images, annotations and camera calibrations provided by KITTI dataset.
\cite{mousavian20173d,qin2019monogrnet,chen2020monopair} try to improve the representation ability of the models by introducing novel geometric constraints.
OFTNet~\cite{roddick2018orthographic} presents an orthographic feature transform to map image-based features into an orthographic 3D space.
MonoDIS~\cite{Simonelli_2019_ICCV} disentangles the loss for 2D/3D detection and jointly trains these two tasks in an end-to-end manner. 
M3D-RPN~\cite{brazil2019m3d} extends the region proposal network (RPN) with 3D box parameters.
These works are orthogonal to our analysis to localization error and the proposed strategies for handling it. 
%Based on that, we explore the underlying reasons behind localization errors, analyze the issues it might bring, and propose our detector.

\noindent
{\bf Monocular 3D detection using additional data.}
To better estimate the 3D bounding boxes, many methods are proposed for effectively using additional data~\cite{Xiang_2015_CVPR,Manhardt_2019_CVPR,chabot2017deep,xu2018multi,cai2020monocular,Ding_2020_CVPR,Godard_2017_CVPR,fu2018deep,wang2019pseudo,Ma_2019_ICCV,Ma_2020_ECCV,brazil2020kinematic}.
Specifically, \cite{Xiang_2015_CVPR,Manhardt_2019_CVPR} use the CAD models as shape templates to get better object geometry.
Deep MANTA~\cite{chabot2017deep}, which takes 3D detection as a key-points detection task, uses more detailed annotated locations of keypoints, \eg wheels, as training labels.
Besides, \cite{xu2018multi,cai2020monocular,Ding_2020_CVPR} estimate the depth maps from off-the-shelf depth estimators~\cite{Godard_2017_CVPR,fu2018deep} trained from larger datasets, and use them to augment the input RGB images.
In addition, \cite{wang2019pseudo,Ma_2019_ICCV} propose to transform the estimated depth maps to pseudo-LiDAR representation, before applying existing LiDAR-based 3D detection designs, and achieve promising performance on KITTI benchmark.
PatchNet~\cite{Ma_2020_ECCV} analyzes the underlying mechanism behind pseudo-LiDAR representation and proposes its corresponding image representation based implementation.
Recently, Kinematic3D~\cite{brazil2020kinematic} propose to use 3D Kalman filter to capture the temporal cues from monocular videos.
In contrast, our method does not use any extra data or annotation, and can still achieve better or competitive performance.

% \begin{figure}[t]
% \begin{center}
% %\fbox{\rule{0pt}{2in} \rule{0.99\linewidth}{0pt}}
% \includegraphics[width=0.85\linewidth]{scale/figs/framework.pdf}
% \end{center}
%   \caption{{\bf Overview.} {\bf (a) Baseline model:} we adopt an one-stage object detection framework to predict seven sub-tasks. Both 2D and 3D bounding boxes can be recovered from the outputs of the model). 
%   }
% \label{fig:model}
% \end{figure}

\noindent
{\bf Misalignment between the definitions of object's center.}
%To cover the position in 3D world space, the center (or other key-points) in image plane are needed. 
To recover the 3D object position, there are two groups of methods. The first group~\cite{zhou2019objects, chen2020monopair, Simonelli_2019_ICCV} use 2D bounding box to obtain 3D position.
In particular, CenterNet \cite{zhou2019objects} regards the center of the 2D bounding box as the projected 3D position in the image plane and back-project it to 3D space with the help of estimated depth and camera parameters.
However, generally speaking, the center of the 2D box and the center of 3D box are not the same.
\cite{chen2020monopair, Simonelli_2019_ICCV} regress an offset to compensate for the difference between them. 
% For the first group of methods, the other two coordinates $(x, y)$ of the 3D position will be influenced by the inaccurate 2D detection results and depth estimation (the main issue in 3D detection). But the coordinates $(x, y)$ of 3D position estimation in our approach will not be influenced by 2D detection or depth.
As the second group, SMOKE~\cite{Liu_2020_CVPR_Workshops} removes the 2D detection and directly estimate 3D position using projected 3D center. This work considers the 2D related sub-tasks are redundant because 2D bounding boxes can be generated from 3D detection results. 
In this work, we revisit this problem and confirm that replacing the 2D center by the projected 3D center can improve the localization accuracy.
Besides, we also find that 2D detection is necessary, because it helps to learn shared features for 3D detection. 
%In summary, we propose joint but separate 2D/3D detection design that can solve the problems of these two groups of designs.
%In this work, we revisit this problem and propose our solution.
%Detailed design choices and analysis will be discussed in  Section~\ref{sec:experment_analysis}.

% \noindent
% {\bf Training samples.}

% When training an object detector, hard samples are particularly valuable as they are more effective to improve the detection performance. How- ever, the random sampling scheme usually results in the selected samples dominated by easy ones. T
% For the hard samples in training p
% OHEM~\cite{OHEM}, focal loss~\cite{Lin_2017_ICCV},
% Libra R-CNN\cite{libra_rcnn}
% Casca \cite{cascade_rcnn,cascade_2010,cascade_wanli}

% \noindent
% {\bf IoU oriented optimization.}
% IoU 
% UnitBox~\cite{unitbox} ...  as loss function in 2D object detection.

% some LiDAR-based approaches~\cite{3diou,3dgiou} apply IoU oriented loss.
% However, 

%%%%   setup
\section{Approach}

\subsection{Problem Definition}
Given are RGB images and the corresponding camera parameters, our goal is to classify and localize the objects of interest in 3D space.
Each object is represented by its category, 2D bounding box ${\bf B_{2D}}$, and 3D bounding box ${\bf B_{3D}}$.
Specifically, ${\bf B_{2D}}$ is represented by its center $\mathbf{c^i}=[x', y']_{2D}$ and size $[h', w']_{2D}$ in the image plane, while ${\bf B_{3D}}$ is defined by its center $[x, y, z]_{3D}$, size $[h, w, l]_{3D}$ and heading angle $\gamma$ in the 3D world space.

\subsection{Baseline Model}
\label{sec:baseline}
\noindent
{\bf Architecture.}
We build our baseline model based on the anchor-free one-stage detector CenterNet~\cite{zhou2019objects}.
Specifically, we use standard DLA-34~\cite{Yu_2018_CVPR} as our backbone for a better speed-accuracy trade-off. 
On top of this, seven lightweight heads (implemented by one $3\times3$ conv layer and one $1 \times 1$ conv layer) are used for 2D detection and 3D detection. More design choices and implementation details can be found in the supplementary material.

\begin{figure}[t]
\centering
%\fbox{\rule{0pt}{2in} \rule{0.99\linewidth}{0pt}}
\includegraphics[width=0.99\linewidth]{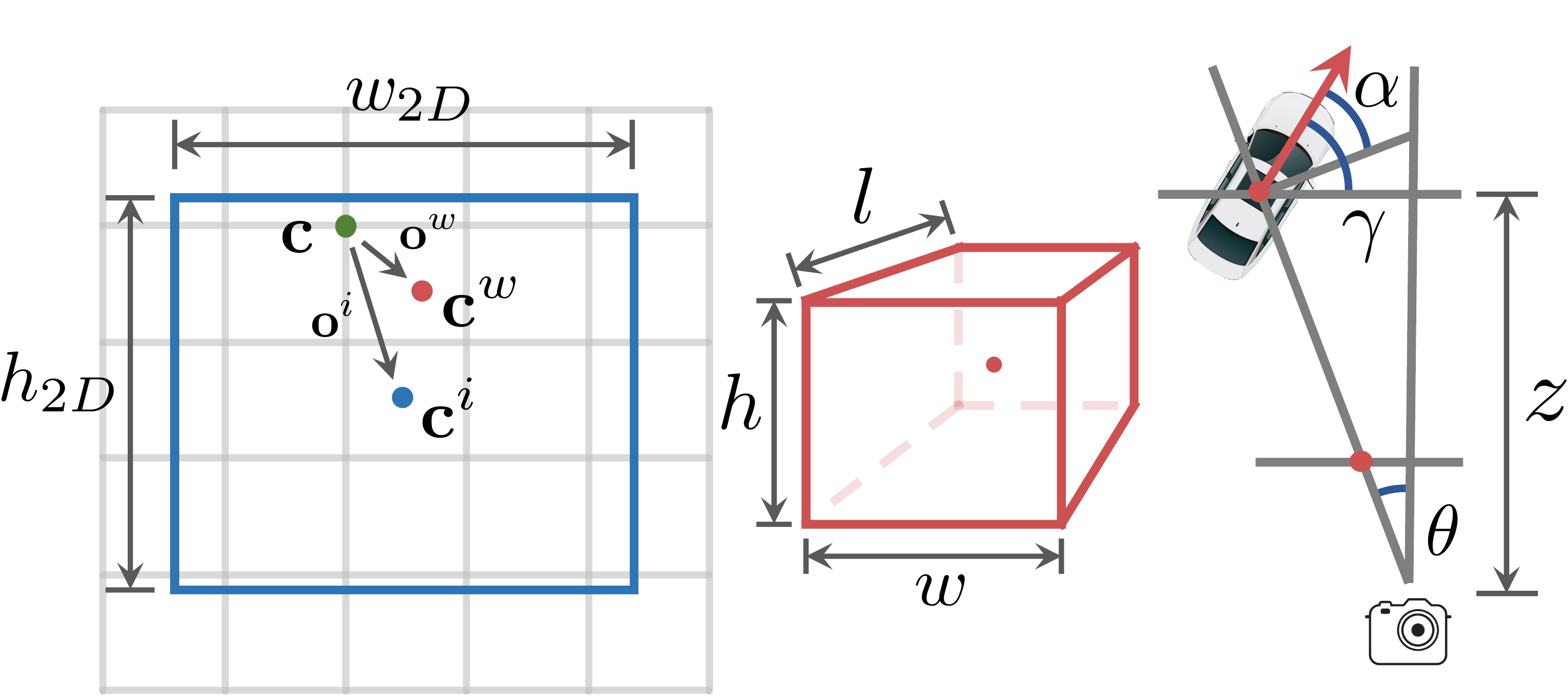}
\caption{{\bf Visualization of the notations} of 2D bounding box in the feature map scale ({\it left}), 3D bounding box in the 3D world space ({\it middle}), and orientation of the object from bird's view ({\it right}). }
\label{fig:geomeric}
\vspace{-8pt}
\end{figure}

\noindent
{\bf 2D detection.}
For 2D detection task, following \cite{Redmon_2016_CVPR,zhou2019objects}, the proposed model outputs a heatmap to indicate the classification score and the coarse center ${\bf c}=(u, v)$ of the object. In existing methods~\cite{zhou2019objects,chen2020monopair,Simonelli_2019_ICCV}, ${\bf c}$ is supervised by the ground-truth 2D bounding box center.
Another branch predict the offset ${\bf o^{i}}=(\Delta u^{i}, \Delta v^{i})$ between the coarse center and the real center of 2D bounding box, and we can get the final 2D box center location ${\bf c^{i}}={\bf c} + {\bf o^{i}}$.
Finally, we use another branch to estimate the size $[w', h']_{2D}$ of 2D bounding box.

\noindent
{\bf 3D detection.}
As for 3D detection, a branch is used for predicting the offset ${\bf o^{w}}=(\Delta u^{w}, \Delta v^{w})$ between the coarse center ${\bf c}$ and the center of projected 3D bounding box ${\bf c^{w}}=[x^w\ y^w]^T={\bf c}+{\bf o^{w}}$. With the known camera intrinsic matrix ${\bf K} \in \mathbb{R}^{3\times 3}$, we can recover the center of object in 3D world space by:
\begin{equation}
\begin{bmatrix}
x \\
y \\
z
\end{bmatrix}_{3D} = {\bf K}^{-1}\begin{bmatrix}
{\bf c^w} \cdot z \\
z
\end{bmatrix}
 = 
{\bf K}^{-1} \cdot
\begin{bmatrix}
x^{w} \cdot z \\
y^{w} \cdot z \\
z
\end{bmatrix}_{2D},
\label{eq:xyz}
\end{equation}
where $z$ is the output of depth branch.
Finally, the last two branches are used to predict the 3D size $[h, w, l]_{3D}$ and orientation $\gamma$, respectively.

% \begin{equation}
% \begin{bmatrix}
% x \\
% y \\
% z
% \end{bmatrix}_{3D} = {\bf K}^{-1}\begin{bmatrix}
% {\bf c} \cdot z \\
% z
% \end{bmatrix}
%  = 
% {\bf K}^{-1} \cdot
% \begin{bmatrix}
% u \cdot z \\
% v \cdot z \\
% z
% \end{bmatrix}_{2D},
% \label{eq:xyz}
% \end{equation}

\noindent
{\bf Losses.}
There are seven loss terms in total, one for foreground/background sample classification, two (center and size) for 2D detection, and four (center, depth, size, and heading angle) for 3D detection.
We adopt the modified Focal Loss used in \cite{Law_2018_ECCV,zhou2019objects} for classification sub-task. 
We use L1 Loss without any anchor for center and size regression in 2D detection task.~For the 3D detection task, uncertainty modeling~\cite{kendall2017uncertainties,cai2020monocular} (see Section~\ref{sec:B} in the supplementary for details) is used for depth estimation; L1 loss is used for 3D center refinement; and multi-bin loss~\cite{Qi_2018_CVPR} (we consider 12 non-overlap equal bins) is used for heading angle estimation.
Lastly, for 3D size estimation, we use L1 loss in baseline (without anchor), and the proposed IoU loss in our model.
The weights for all loss items are set to 1.

\subsection{Error Analysis}

\begin{table}[!t]
\begin{center}
\begin{tabular}{lc|lc}
\toprule
baseline & 11.12 & ground truth & 99.97 \\
w/ gt proj. center & 23.90 & w/o gt proj. center & 46.33\\
w/ gt depth & 38.01 & w/o gt depth & 25.25 \\
w/ gt 3D location &  78.84 & w/o gt 3D location & 12.13  \\
w/ gt 3D size &  11.96 &  w/o gt 3D size & 80.50  \\
w/ gt orientation & 11.88 &  w/o gt orientation & 70.89  \\

\bottomrule
\end{tabular}
\end{center}
\vspace{-10pt}
\caption{{\bf Error analysis.}
{\it Left:} We replace the outputs of 3D detection related branches with the ground truth values.
{\it Right:} We replace the values of ground truth with the predicted results.
Metric is ${\rm AP}_{40}$ for 3D detection under moderate setting on the KITTI val set. `proj. center' denotes  the projected 3D center $\mathbf{c^w}$ on the image plane. }
\label{table:erroranalysis}
\vspace{-7pt}
\end{table}

In this section, we explore what restricts the performance of monocular 3D detection. Inspired by CenterNet~\cite{zhou2019objects} and CornerNet~\cite{Law_2018_ECCV} in the 2D detection field, we conduct an error analysis for different prediction items on KITTI \emph{validation} set via replacing each predictions with ground truth value and evaluating the performance. 
Specifically, we replace each output head with its ground truth according to the practice of \cite{Law_2018_ECCV,zhou2019objects}. 
As shown in Table~\ref{table:erroranalysis}, if we replace projected 3D center ${\bf c^{w}}$ predicted from baseline model with its ground-truth, the accuracy is improved from 11.12\% to 18.97\%. On the other hand, depth can improve the accuracy to $38.01\%$. 
%Therefore, depth is the most important single factor that influences the 3D detection accuracy. 
If we consider both depth and projected center, \ie replacing the predicted 3D locations $[x, y, z]_{3D}$ with ground-truth results, then the most obvious improvement is observed. Therefore, the low accuracy of monocular 3D detection is mainly caused by localization error. On the other hand, 
according to Equation~\ref{eq:xyz}, depth estimation and center localization jointly determine the position of the object in 3D world space.
Compared with the ill-posed depth estimation from a monocular image, improving the accuracy of center detection is a more feasible way.

\begin{figure}[t]
\begin{center}
% \fbox{\rule{0pt}{1.4in} \rule{0.45\linewidth}{0pt}}
% \fbox{\rule{0pt}{1.4in} \rule{0.45\linewidth}{0pt}}
\includegraphics[width=0.99\linewidth]{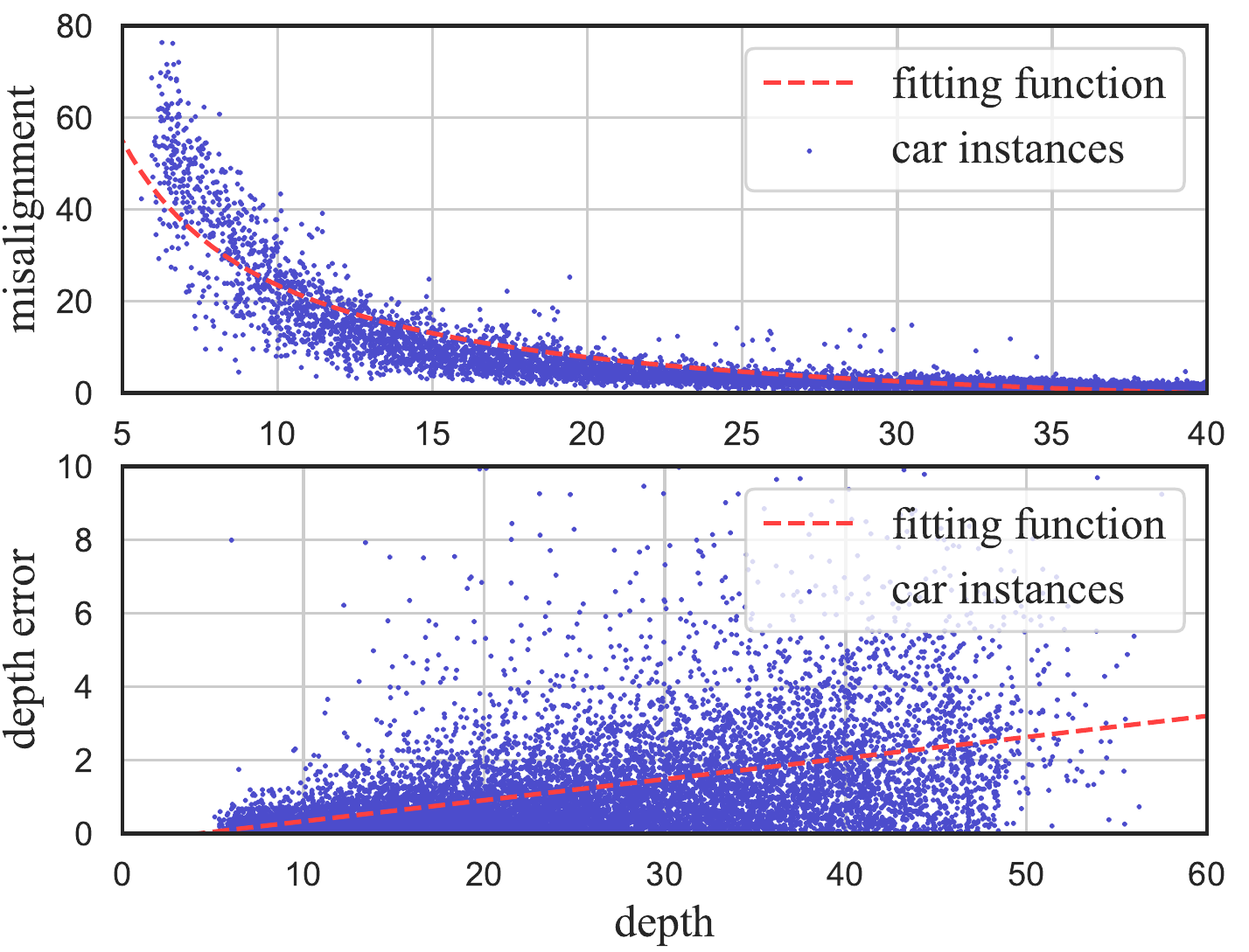}
\end{center}
\vspace{-15pt}
\caption{{\bf Statistics.} 
{\it Top}: the misalignment (in pixel, collected on the KITTI \emph{trainval} set under moderate setting) between the center of 2D bounding box and the projected 3D center in the image plane.
{\it Bottom}: the depth errors (in meter, trained on the KITTI \emph{training} set, tested on the \emph{validation} set)
These two statistics are presented as the function of the depth (x-axis).
}
\label{fig:stats}
\end{figure}

Table~\ref{table:analysis_loc_error} shows localization errors introduced by inaccurate center detection.
Furthermore, the mean shape of cars in KITTI dataset is $[1.53m, 1.63m, 3.53m]$ for $[h, w, l]_{3D}$.
Suppose that all other quantities are correct and the localization error is aligned with the length $l$ (resulting in the maximum tolerance), the IoU can be computed by:
\begin{equation}
IoU = \frac{3.53 - \Delta_{loc}}{3.53 + \Delta_{loc}},
\end{equation}
where $\Delta_{loc}$ represents the localization error.
According to the official setting, the IoU threshold should be set to 0.7, thus the theoretically acceptable maximum error is $0.62m$.
However, an error of only 4-8 pixels in the image (1-2 pixel in $4\times$ down sampling feature map) will cause the object at 60 meters cannot be detected correctly.
Coupled with the errors accumulated by other tasks such as depth estimation (Figure~\ref{fig:stats} shows the errors of depth estimation), it becomes an almost impossible task to accurately estimate the 3D bounding box of distant objects from a single monocular image, unless the depth estimation is accurate enough (not achieved to date).
%In this case, does the network still need to learn how to fit these samples?

\begin{table}[!t]
\begin{center}
\begin{tabular}{cc|ccccc}
\toprule
${\Delta u}$ & ${\Delta v}$  & 5m & 10m & 20m & 40m & 60m\\ 
\hline
2 & 2 & 0.02 & 0.04 & 0.08 & 0.16 & 0.24  \\
4 & 2 & 0.03 & 0.06 & 0.13 & 0.25 & 0.38  \\
6 & 2 & 0.04 & 0.09 & 0.18 & 0.36 & 0.54  \\
6 & 4 & 0.05 & 0.10 & 0.20 & 0.41 & 0.61  \\
8 & 2 & 0.06 & 0.12 & 0.23 & 0.47 & 0.70  \\
8 & 6 & 0.07 & 0.14 & 0.28 & 0.57 & 0.85  \\
\bottomrule
\end{tabular}
\end{center}
\vspace{-10pt}
\caption{{\bf Localization error} (in meter) caused by center shifting in image plane (in pixel).}
\label{table:analysis_loc_error}
\vspace{-10pt}

\end{table}

To better show the importance of center localization, we show the localization error in 3D space caused by shifting the center in image plane in Table~\ref{table:analysis_loc_error}.

%First, we calculate the distance in image plane between each predicted object and each object from ground truth in the same picture as:
%\begin{equation}
%	D_{i, j} = ||c^i - c^j||_1 = \sqrt{({x^i}'-{x^j}')^2+{y^i}'-{y^j}')^2},
%\end{equation}
%where $i \in I$ and $j \in J$ represent the indices for predicted object and object from ground truth, respectively; $I$ and $J$ represent all predicted object and ground truth object in the image; $D_{i, j}$ denotes the distance in image plane between predicted object $i$ and object $j$ from ground truth. We further find the ground truth object $j$ closest to the predicted object $i$ in 2D image plane as:
%\begin{equation}
%    j_i' = \argmin\limits_{{i \in I}} D_{i, j},
%\end{equation}
%where $j_i'$ indicates the index of ground truth object closest to the predicted object $i$.
%We replace the prediction of predicted object $i$ with the ground truth value of object $j_i'$ and evaluate the performance on KITTI \emph{validation} set.

\subsection{Revisiting Center Detection}
\label{sec:Keypoint}

\begin{figure}[t]
\centering
%\fbox{\rule{0pt}{1.6in} \rule{0.99\linewidth}{0pt}}
\includegraphics[width=0.8\linewidth]{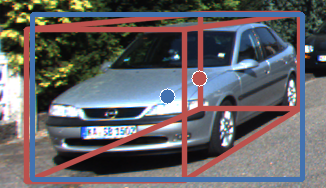}
\caption{{\bf Visualization of the misalignment} between the center of the 2D bounding box (\textcolor{blue}{blue}) and the projected 3D center (\textcolor{red}{red}) in image plane.}
%\vspace{-5}
\label{fig:misalignment}
\end{figure}

\noindent
\textbf{Our design for center detection}. For estimating the coarse center ${\bf c}$, our design is simple. In particular,  we 1) use the projected 3D center $\mathbf{c^w}$ as the ground-truth for the branch estimating coarse center ${\bf c}$ and 2) force our model to learn features from 2D detection simultaneously. 
This simple design is from our analysis below.

\noindent
\textbf{Analysis 1.}
As shown in Figure~\ref{fig:misalignment}, there is a misalignment between the 2D bounding box center $\mathbf{c^i}$ and the projected center $\mathbf{c^w}$  of the 3D bounding box. According to the formulation in Equation~\ref{eq:xyz}, the projected 3D center $\mathbf{c^w}$ should be the key for recovering the 3D object center $[x, y, z]_{3D}$.
The key problem here is what should be the supervision for the coarse center ${\bf c}$. 
%There is no doubt that ${\bf o^{i}}$ and ${\bf o^{w}}$ should be trained under the supervision of (${\bf c^{i}-c}$) and (${\bf c^{w}-c})$.
Some works~\cite{chen2020monopair,Simonelli_2019_ICCV} choose to use 2D box center ${\bf c^{i}}$ as its label, which is not related to the 3D object center, making the estimation of the coarse center not aware of the 3D geometry of the object. 
Here we choose to adopt the projected 3D center $\mathbf{c^w}$ as the ground-truth for the coarse center ${\bf c}$. This helps the branch for estimating the coarse center aware of 3D geometry and more related to the task of estimating 3D object center, which is the key of localization problem (see Section~\ref{sec:E} in supplementary materials for visualizations).

\noindent
\textbf{Analysis 2.}
Note that SMOKE~\cite{Liu_2020_CVPR_Workshops} also use the projected 3D center ${\bf c^{w}}$ as the label of the coarse center ${\bf c}$. However, they discard 2D detection related branches while we preserve them. In our design, the coarse center ${\bf c}$ supervised by the projected 3D center $\mathbf{c^w}$ is also used for estimating the 2D bounding box center $\mathbf{c^i}$. With our design, we force a 2D detection branch to estimate an offset ${\bf o^{i}}={\bf c^{i}}-{\bf c}$ between the real 2D center and the coarse 2D center. This makes our model \emph{aware of the geometric information of the object}. 
Besides, another branch is used to estimate the size of the 2D bounding box so that \emph{the shared features can learn some cues that benefit to depth estimation due to the perspective projection.}
In this way, the 2D detection serves as an auxiliary task that helps to learn better 3D aware features.
%Here we build a bridge to compare the two options. Specifically, we use the first one to train our baseline model, and conduct the control experiment by using ${\bf c^{w}}$ as the ground-truth for ${\bf c}$. To eliminate the influence of code implementation, we implement all variant models based on the baseline we use in this work.

\subsection{Training Samples}
\label{sec:sample}
Different from ~\cite{OHEM,Lin_2017_ICCV} which force network focus on the `hard' samples, we argue that ignoring some extremely `hard' cases can improve the overall performance for the monocular 3D detection task.
Both the results shown in Figure~\ref{fig:evalbyrange} and the analysis conducted in Section~\ref{sec:experment_analysis} illustrate there is a strong relationship between the distance of the object and the difficulty of detecting it.
According to this, two schemes are proposed on how to generate the  object-level training weight $w_{i}$ for sample $i$.

\noindent
\textbf{Scheme 1, hard coding.}
This scheme discard all samples over a certain distance:
\begin{equation}
w_{i}=\left\{
\begin{aligned}
1 \quad \mbox{if} \ \ d_{i} \leq s \\
0 \quad \mbox{if} \ \ d_{i}  > s 
\end{aligned}
\right.
\end{equation}
where $d_{i}$ denotes the depth of sample $i$, and $s$ is the threshold of depth which is set to 60 meters in our implementation. In this way, the samples with depth larger than $s$ will not be used in the training phase.

\noindent
\textbf{Scheme 2, soft coding.}
The other one is soft encoding, and we generate it using a reverse sigmoid-like function:
\begin{equation}
w_{i} = \frac{1}{1+e^{(d_{i}-c)/T}},
\end{equation}
where $c$ and $T$ are the hyper-parameters to adjust the center of symmetry and bending degree, respectively.
When $c=s$ and $T \to 0 $, it is equivalent to the hard encoding scheme. When $T \to \infty$, it is equivalent to using the same weight for all samples.
By default, $c$ and $T$ are set to 60 and 1, and the empirical experiments in Section~\ref{sec:experiment} find that scheme 1 and scheme 2 are both effective and have similar results.

\subsection{IoU Oriented Optimization}
\label{sec:IoU}
% \begin{figure}[t]
% \begin{center}
% \fbox{\rule{0pt}{1.6in} \rule{0.99\linewidth}{0pt}}
%   %\includegraphics[width=0.8\linewidth]{egfigure.eps}
% \end{center}
%   \caption{This fig will show the dimension aware loss.}
% \label{fig:dimension_aware}
% \end{figure}

Recently, some LiDAR based 3D detectors~\cite{3dgiou,3diou} applied the IoU oriented optimization~\cite{rezatofighi2019generalized}.
However, determining the 3D center of object is an very challenging task for monocular 3D detection, and the localization error often reaches several meters (see Section~\ref{sec:experment_analysis}).
In this case, localization related sub-tasks (such as depth estimation) will overwhelm others (such as 3D size estimation), if we apply IoU based loss function directly.
Moreover, depth estimation from monocular image itself an ill-posed problem, and this kind of contradiction will make the training process collapse.
Disentangling each loss item and optimize them independently is a another choice~\cite{Simonelli_2019_ICCV}, but this ignores the correlation of each component to the final result.
To alleviate this problem, we propose a IoU oriented optimization for 3D size estimation.
Specifically, suppose all prediction items except the 3D size ${\bf s}=[h, w, l]_{3D}$ are completely correct, then we can get (details for deriving can be found in supplementary materials):
\begin{equation}
\frac{\partial IoU}{\partial h} : \frac{\partial IoU}{\partial w} : \frac{\partial IoU}{\partial l}
 \approx \frac{1}{h} : \frac{1}{w} : \frac{1}{l}.
 \label{eq:iou}
\end{equation}
Accordingly, we can adjust the weight of each side by its partial derivative \wrt IoU (in magnitude), and the loss function of the 3D size estimation can be modified to:
\begin{equation}
\mathcal{L}_{size} = || \frac{({\bf s} - {\bf s^{*})}}{{\bf s}} ||_{1}, 
\label{eq:lsize}
\end{equation}
where $|| \cdot ||_{1}$ represent the $L_1$ norm.
Note that, compared with the standard 3D size loss $\mathcal{L}_{size}'=||{\bf s} - {\bf s^{*}}||_{1}$ used in the baseline model, our new loss's magnitude is changed.
To compensate it, we compute $\mathcal{L}_{size}'$ once more, and dynamically generate the compensate weight $w_{s}=|\mathcal{L}_{size}'/\mathcal{L}_{size}|$, so that the mean value of the final loss function $w_{s} \cdot \mathcal{L}_{size}$ is equal to the standard one.
By this way, the proposed loss can be regard as a re-distribution of the standard L1 loss.

\begin{table*}[!t]
\centering
\resizebox{\linewidth}{!}{
\begin{tabular}{l||c|ccc|ccc|ccc|c}
\toprule
\multirow{2}{*}{Method} & \multirow{2}{*}{Extra data} & \multicolumn{3}{c|}{3D } & \multicolumn{3}{c|}{BEV}  & \multicolumn{3}{c|}{AOS} & \multirow{2}{*}{Runtime}\\ 
\cline{3-11} 
 ~ & ~ & Easy & Mod. & Hard  & Easy & Mod. & Hard & Easy & Mod. & Hard & ~\\ 
\hline
Decoupled-3D~\cite{cai2020monocular} & Yes
& 11.08 & 7.02  & 5.63  
& 23.16 & 14.82 & 11.25 
& 87.34 & 67.23 & 53.84 & - \\
AM3D~\cite{Ma_2019_ICCV} & Yes
& 16.50 & 10.74 & 9.52  
& 25.03 & 17.32 & 14.91
&-  &-  &- & $\sim$400 ms \\
PatchNet~\cite{Ma_2020_ECCV} & Yes
&15.68 & 11.12 & 10.17  
&22.97 & 16.86 & 14.97
&-  &-  &- & $\sim$400 ms \\
D4LCN~\cite{Ding_2020_CVPR}  & Yes
& 16.65 & 11.72 & 9.51  
& 22.51 & 16.02 & 12.55
& 90.01 & 82.08 & 63.98 & -\\
Kinematic3D~\cite{brazil2020kinematic}  & Yes
& 19.07 & 12.72 & 9.17  
& 26.69 & 17.52 & 13.10
& 58.33 & 45.50 & 34.81 & 120ms\\

\hline
GS3D~\cite{Li_2019_CVPR} & No 
& 4.47  & 2.90  & 2.47 
& 8.41  & 6.08  & 4.94 
& 85.79 & 75.63 & 61.85 & $\sim$2000 ms \\
MonoGRNet~\cite{qin2019monogrnet} & No 
& 9.61  & 5.74  & 4.25 
& 18.19 & 11.17 & 8.73 
& - & - & - & 60 ms \\ 
MonoDIS~\cite{Simonelli_2019_ICCV} & No  
& 10.37 & 7.94  & 6.40 
& 17.23 & 13.19 & 11.12 
& - & - & - & - \\  
M3D-RPN~\cite{brazil2019m3d} & No 
& \underline{14.76} & 9.71  & 7.42 
& \underline{21.02} & 13.67 & 10.23 
& 88.38 & 82.81 & 67.08 & 161 ms\\ 
SMOKE~\cite{Liu_2020_CVPR_Workshops} & No 
& 14.03 & 9.76 & 7.84
& 20.83 & 14.49 & 12.75 
& \underline{92.94} & \underline{87.02} & \underline{77.12} & {\bf 30 ms} \\ 
MonoPair~\cite{chen2020monopair} & No 
& 13.04 & \underline{9.99} & \underline{8.65} 
& 19.28 & \underline{14.83} & \underline{12.89} 
& 91.65 & 86.11 & 76.45 & 57 ms\\ 
Ours & No  
& {\bf 17.23} & {\bf 12.26} & {\bf 10.29} 
& {\bf 24.79} & {\bf 18.89} & {\bf 16.00} 
& {\bf 93.46} & {\bf 90.23} & {\bf 80.11} & \underline{40 ms}\\  
\hline
Improvement  & -
& +2.47 & +2.27 & +1.64 
& +3.77 & +4.06 & +3.11 
& +0.52 & +3.21 & +2.99 & - \\  
\bottomrule
\end{tabular}}
\vspace{-4pt}
\caption{{\bf Performance of the Car category on the KITTI \emph{test} set.} 
Methods are ranked by moderate setting (same as KITTI leaderboard). 
We highlight the best results in {\bf bold} and the second place in \underline{underlined}.}
\label{table:testset}
\vspace{-2pt}
\end{table*}

\subsection{Implementation}
\noindent
{\bf Training.}
We train our model on two GTX 1080Ti GPUs with a batch size of 16 in an end-to-end manner for 140 epochs.
We use Adam optimizer with initial learning rate $1.25e^{-3}$, and decay it by ten times at 90 and 120 epochs.
The weight decay is set to $1e^{-5}$ and the warmup strategy is also used for the first 5 epochs.
To avoid over-fitting, we adopt the random cropping/scaling (for 2D detection only) and random horizontal flipping.
Under this setting, it takes around 9 hours for whole training process.
%on our environment.

\noindent
{\bf Inference.}
During the inference phase, we obtain the prediction results from the parallel decoders.
To decoding the results, similar to \cite{zhou2019objects}, we conduct the efficient non-maxima suppression (NMS) on center detection results using a $3 \times 3$ max pooling kernel.
Then, we recover 2D/3D bounding boxes according to encoding strategy introduced in Section~\ref{sec:baseline} and use the score of center detection as the confidence of predicted results.
Finally, we discard predictions with confidence less than 0.2.

%%%%   experiments

\begin{table*}[!t]
\centering
\resizebox{\linewidth}{!}{
\begin{tabular}{l||ccc|ccc|ccc|ccc}
\toprule
\multirow{2}{*}{Method} & \multicolumn{3}{c|}{3D@IOU=0.7} & \multicolumn{3}{c|}{BEV@IOU=0.7} & \multicolumn{3}{c|}{3D@IOU=0.5} & \multicolumn{3}{c}{BEV@IOU=0.5}\\ 
\cline{2-13} 
 ~ & Easy & Mod. & Hard  & Easy & Mod. & Hard & Easy & Mod. & Hard & Easy & Mod. & Hard \\ 
\hline
CenterNet~\cite{zhou2019objects} 
& 0.60 & 0.66 & 0.77 
& 3.46 & 3.31 & 3.21 
& 20.00 & 17.50 & 15.57
& 34.36 & 27.91 & 24.65 \\  
MonoGRNet~\cite{qin2019monogrnet} 
& 11.90 & 7.56  & 5.76 
& 19.72 & 12.81 & 10.15 
& 47.59 & 32.28 & 25.50
& 48.53 & 35.94 & 28.59 \\  
MonoDIS~\cite{Simonelli_2019_ICCV}  
& 11.06 & 7.60 & 6.37 
& 18.45 & 12.58 & 10.66 
& - & - &
& - & - &\\  
M3D-RPN~\cite{brazil2019m3d}
& 14.53 & 11.07 & 8.65 
& 20.85 & 15.62 & 11.88 
& 48.53 & 35.94 & 28.59
& 53.35 & 39.60 & 31.76\\ 
MonoPair~\cite{chen2020monopair}
& \underline{16.28} & \underline{12.30} & \underline{10.42} 
& \underline{24.12} & \underline{18.17} & \underline{15.76} 
& \underline{55.38} & \underline{42.39} & {\bf37.99}
& {\bf 61.06} & {\bf 47.63} & {\bf 41.92}\\ 
Ours   
& {\bf 17.45} & {\bf 13.66} & {\bf 11.68} 
& {\bf 24.97} & {\bf 19.33} & {\bf 17.01} 
& {\bf 55.41} & {\bf 43.42} & \underline{37.81} 
& \underline{60.73} & \underline{46.87} & \underline{41.89}\\  
\hline
Improvement   
& +1.17 & +1.36 & +1.26
& +0.85 & +1.16 & +1.25
& +0.03 & +1.03 & -0.18
& -0.33 & -0.80 & -0.03 \\  
\bottomrule
\end{tabular}}
\vspace{-4pt}
\caption{{\bf Performance of the Car category on the KITTI \emph{validation} set.} 
Methods are ranked by moderate setting (same as KITTI leaderboard). 
We highlight the best results in {\bf bold} and the second place in \underline{underlined}.}
\label{table:valset}
\vspace{-8pt}
\end{table*}

\section{Experimental Results}
\label{sec:experiment}

\subsection{Setup}
\noindent
\textbf{Dataset.}
We evaluate our method on the challenging KITTI dataset~\cite{Geiger2013IJRR,geiger2012we}, which provides 7,481 images for training and 7,518 images for testing.
% 3D detection and bird's view detection are evaluated in three different subsets: easy, moderate and hard, according to the occlusion and truncation levels of objects. 
Since the ground truth for the test set is not available and the access to the test server is limited, we follow the protocol of prior works~\cite{Chen_2016_CVPR,chen20153d,Chen_2017_CVPR} to divide the training data into a training set (3,712 images) and a validation set (3,769 images). 
We conduct ablation studies based on this split and also report final results which trained on all 7,481 images and tested by KITTI official server.

% Due to space limitations, we only report the Car detection results of monocular images in the main paper. 
% More results about stereo pairs and Pedestrian/Cyclist can be found in Appendix.

\noindent
\textbf{Metrics.}
The KITTI dataset provides many widely used benchmarks for autonomous driving scenarios, including 3D detection, bird's eye view (BEV) detection, and average orientation similarity (AOS).
We report the Average Precision with 40 recall positions (${\rm AP}_{40}$) \cite{Simonelli_2019_ICCV} under three difficultly settings (easy, moderate, and hard) for those tasks.
We mainly focus on the {\bf Car} category, and also report the performances of the {\bf Pedestrian} and {\bf Cyclist} categories for reference.
The default IoU threshold are 0.7, 0.5, 0.5 for these categories.

\subsection{Main Results}

\noindent
{\bf Results on the KITTI \emph{test} set.}
As shown in Table~\ref{table:testset}, we report our results of the {\bf Car} category on KITTI \emph{test} set. 
Overall, our method achieves superior results over previous methods across all settings under fair conditions.
For instance, the proposed method obtains {\bf 2.47/2.27/1.64} improvements under easy/moderate/hard setting for 3D detection task.
Besides, our method achieves {\bf 18.89/90.23} in BEV detection/AOS task under moderate setting, improving previous best results by {\bf 4.06/4.12} ${\rm AP}_{40}$.
Compared with the methods with extra data, the proposed method still get comparable performances, which further proves the effectiveness of our model.

\noindent
{\bf Results on the KITTI \emph{validation} set.}
We also present our model's performance on the KITTI \emph{validation} set in Table~\ref{table:valset}.
Note that some methods directly use the pre-trained model provided by DORN~\cite{fu2018deep} as their depth estimator.
However, the DORN's training set overlaps with the validation set of KITTI 3D, so we are not comparing these methods here.
We can find that the proposed model performs better than all previous methods in 3D detection task.
For BEV detection task, our method outperforms all methods except for MonoPair.
Compared with MonoPair, our method is better at detecting objects under strict conditions (0.7 IoU threshold), while MonoPair is slightly better at catching samples under loose conditions (0.5 IoU threshold).
Also note that our method shows better performance consistency between the validation set and test set. 
This indicates that our method has better generalization ability, which is of great significance in autonomous/assisted driving.
% Note that the contribution of MonoPair in modeling the geometric relationship between adjacent objects is complementary to ours. Therefore, the method used in MonoPair can be naturally adopted in our approach for further boosting accuracy.

\begin{table}[!t]
\centering
\resizebox{\linewidth}{!}{
\begin{tabular}{l||c|ccc}
\toprule
Method & Cat.  & Easy & Mod. & Hard \\
\hline
M3D-RPN~\cite{brazil2019m3d} & Ped. &
5.65 / 4.92 & 4.05 / 3.48 & 3.29 / 2.94 \\
MonoPair~\cite{chen2020monopair} & Ped. & 
10.99 / 10.02 & 7.04 / 6.68 & 6.29 / 5.53 \\
Ours & Ped.  & 
10.73 / 9.64 & 6.96 / 6.55 & 6.20 / 5.44 \\
\hline
M3D-RPN~\cite{brazil2019m3d} & Cyc. & 
1.25 / 0.94 & 0.81 / 0.65 & 0.78 / 0.47 \\
MonoPair~\cite{chen2020monopair} & Cyc.  & 
4.76 / 3.79 & 2.87 / 2.12 & 2.42 / 1.83 \\
Ours & Cyc.  & 
5.34 / 4.59 & 3.28 / 2.66 & 2.83 / 2.45 \\
\bottomrule
\end{tabular}}
\vspace{-3pt}
\caption{{\bf Benchmark for Pedestrian/Cyclist detection on the KITTI \emph{test} set.} 
Metric is ${\rm AP}_{40}$ for BEV/3D detection task at 0.5 IoU threshold.}
\label{table:pedcyc}
\vspace{-5pt}
\end{table}

\begin{table}[t]
\centering
\resizebox{\linewidth}{!}{
\begin{tabular}{l||ccc}
\toprule
  & Easy & Mod. & Hard  \\ 
\hline
baseline
& 20.29 / 14.51 & 16.15 / 11.12 & 14.07 / 9.97\\
 + p.
& 23.10 / 15.78 & 18.15 / 12.65 & 16.11 / 10.62\\
 + p.+I.
& 23.89 / 16.12 & 18.34 / 12.97 & 16.69 / 10.99\\
 + p.+I.+s.
& 24.97 / 17.45 & 19.33 / 13.66 & 17.01 / 11.68 \\
\bottomrule
\end{tabular}}
\vspace{-3pt}
\caption{{\bf Results on accumulating the proposed approaches on the KITTI \emph{validation} set.} Metric is ${\rm AP}_{40}$ of the Car category for BEV/3D detection. `p.' denotes using projected 3D center for supervising the coarse center. `I.' denotes using our IoU loss design. `s.' denotes the design for discarding distant samples. }
\label{table:ablation}
\vspace{-10pt}
\end{table}

\noindent
{\bf Latency analysis.}
We test the proposed model on a single GTX 1080Ti GPU with a batchsize of 1 for runtime analysis.
As shown in Table~\ref{table:testset}, the proposed method can run at 25 FPS, meeting the requirement of real-time detection.  
Specifically, our method runs $4 \times$ faster than the two-stage detector M3D-RPN.
Compared with MonoPair, which shares a similar framework as ours, our method can still save 16 ms for one image in the inference phase, mainly because: 
1) we use standard DLA-34 as our backbone, instead of modified DL4-34 with DCN~\cite{dai2017deformable,Zhu_2019_CVPR}.
2) we apply fewer prediction heads in our model.
3) we don't need any post-processing.
SMOKE~\cite{Liu_2020_CVPR_Workshops} can run faster than our method.
However, it only conducts 3D detection while the proposed method can perform 2D detection and 3D detection jointly.

Besides, although the detectors with pretrained depth estimator usually have promising performance, the additional depth estimator introduce lots of computational overheads (\eg the most commonly used DORN~\cite{fu2018deep} takes about 400 ms to process a standard KITTI image. See \hyperlink{http://www.cvlibs.net/datasets/kitti/eval_depth.php?benchmark=depth_prediction}{KITTI Depth Benchmark} for more details).

\begin{table}[!t]
\centering
\resizebox{\linewidth}{!}{
\begin{tabular}{c||ccc|ccc}
\toprule
~ & PC & RF & MT & Easy & Mod. & Hard\\ 
\hline
{\bf a} & - & - & \checkmark & 
98.08 / 1.32 & 92.31 / 1.04 & 84.75 / 1.16 \\
{\bf b} &- & \checkmark & \checkmark & 
98.08 / 13.98 & 92.31 / 10.81 & 84.75 / 9.59\\
{\bf c} & \checkmark & \checkmark & - & 
94.55 / 12.31 & 88.79 / 10.30  & 79.29 / 8.82   \\
{\bf d} & \checkmark & \checkmark & \checkmark & 
98.42 / 16.08 & 92.74 / 13.04 & 83.04 / 11.16  \\
\bottomrule
\end{tabular}}
\vspace{-3pt}
\caption{{\bf Analysis for center definition and multitask learning.} 
Metrics are ${\rm AP}_{40}$ of the Car category for 2D/3D detection tasks.
`PC', `RF', and `MT' represent `projected 3D center', `refinement', and `multi-task learning'.}
\label{table:analysis_center}
\vspace{-10pt}
\end{table}

\subsection{Pedestrian/Cyclist Detection}
Here we present the Pedestrian/Cyclist detection results on the KITTI \emph{test} set in Table~\ref{table:pedcyc}.
Compared with cars, pedestrians/cyclists are more difficult to detect, and only \cite{brazil2019m3d,chen2020monopair} provide the performances of those categories on KITTI test set.
Specifically, the proposed method performs better than \cite{brazil2019m3d} and gets comparable results with  \cite{chen2020monopair}.
But it is important to note that, since the number of training samples for those two categories is quite small, the performance may fluctuate to some extent.

\subsection{Analysis}
\label{sec:experment_analysis}

\noindent
{\bf Accumulation of the proposed designs.}
Table~\ref{table:ablation} shows experimental results evaluating how the proposed designs contribute to the overall performance for this task. 
Our design in Section~\ref{sec:Keypoint}, which uses projected 3D center for supervising center detection and influencing 2D detection (`+p.' in Table~\ref{table:ablation}), improves 3D detection accuracy by 1.5. The IoU loss design in Section~\ref{sec:IoU} further improves the accuracy by 0.3. And the design for discarding distant samples in Section~\ref{sec:sample} leads to 0.7 improvement.

\begin{figure*}[t]
\centering
\includegraphics[width=0.310\linewidth]{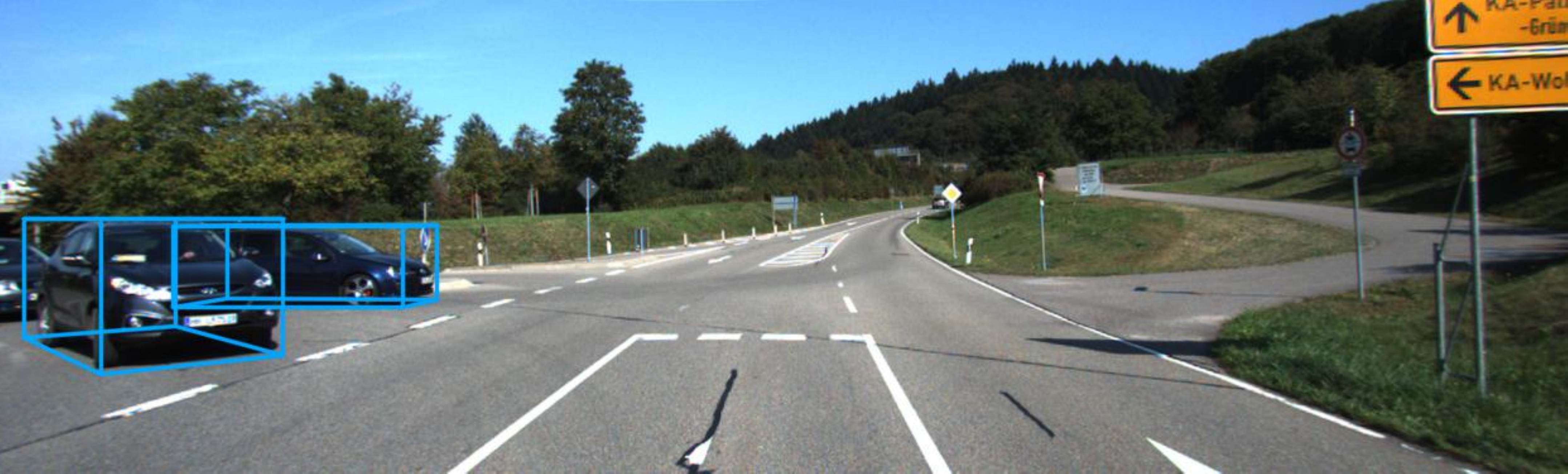}
\includegraphics[width=0.310\linewidth]{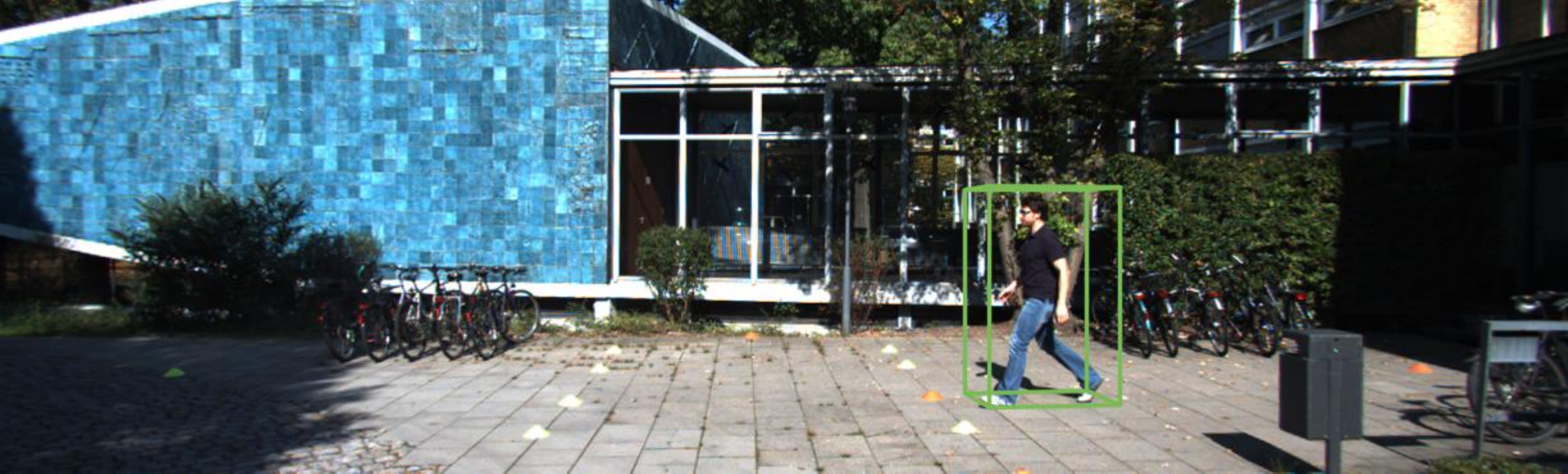}
\includegraphics[width=0.310\linewidth]{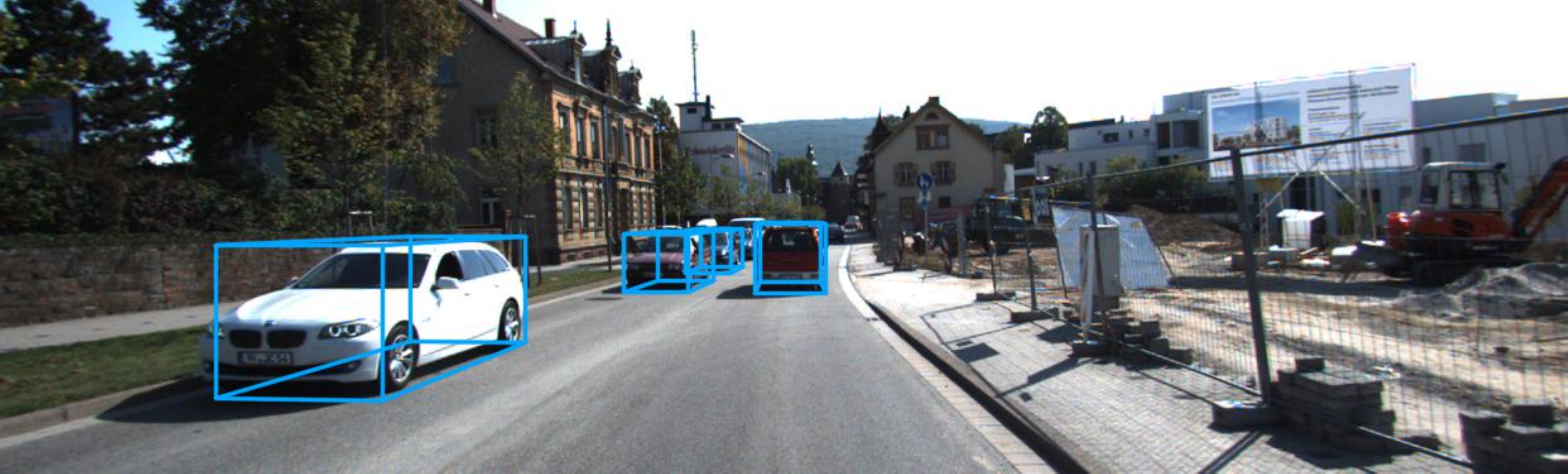}  \\
\includegraphics[width=0.310\linewidth]{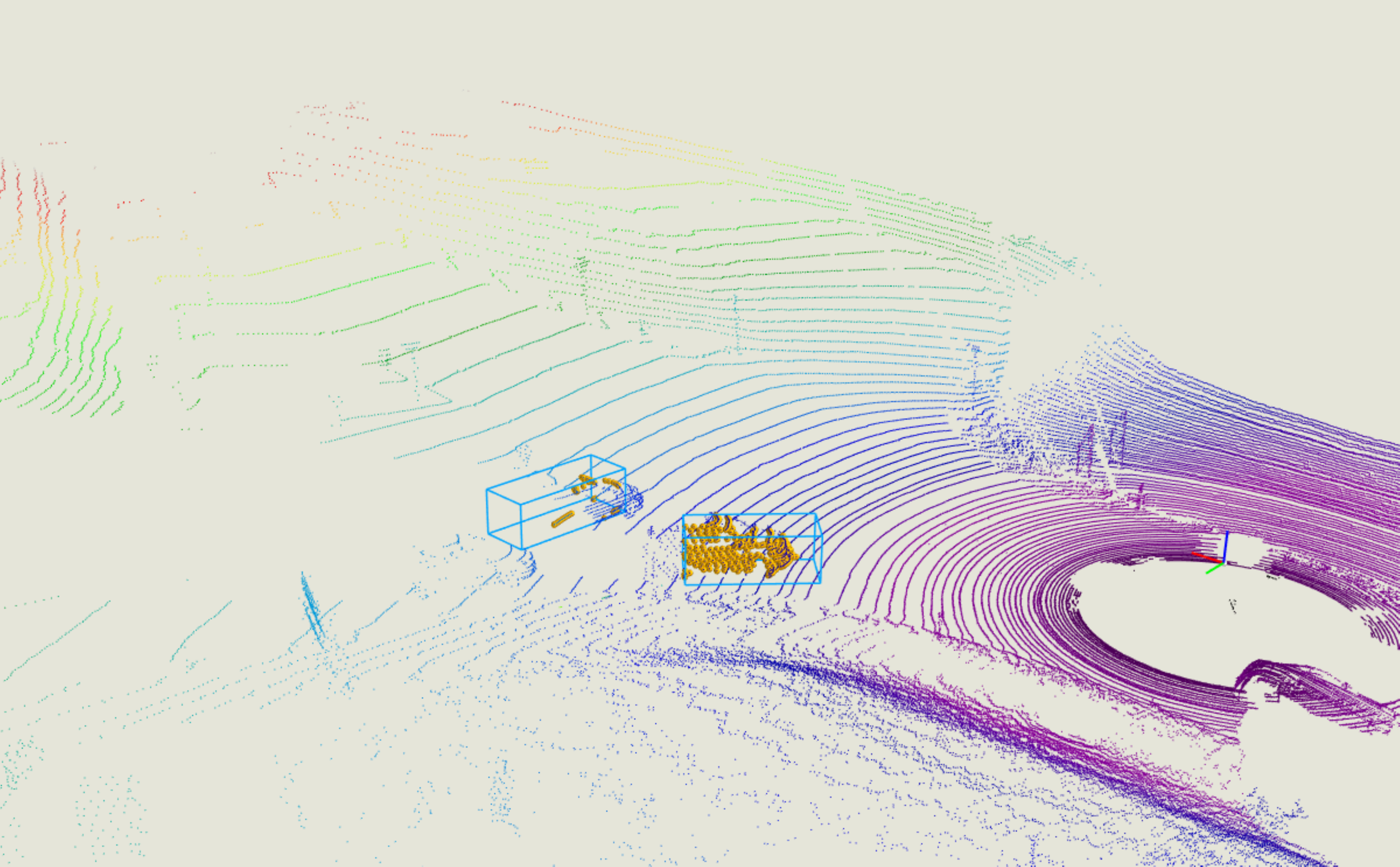}
\includegraphics[width=0.310\linewidth]{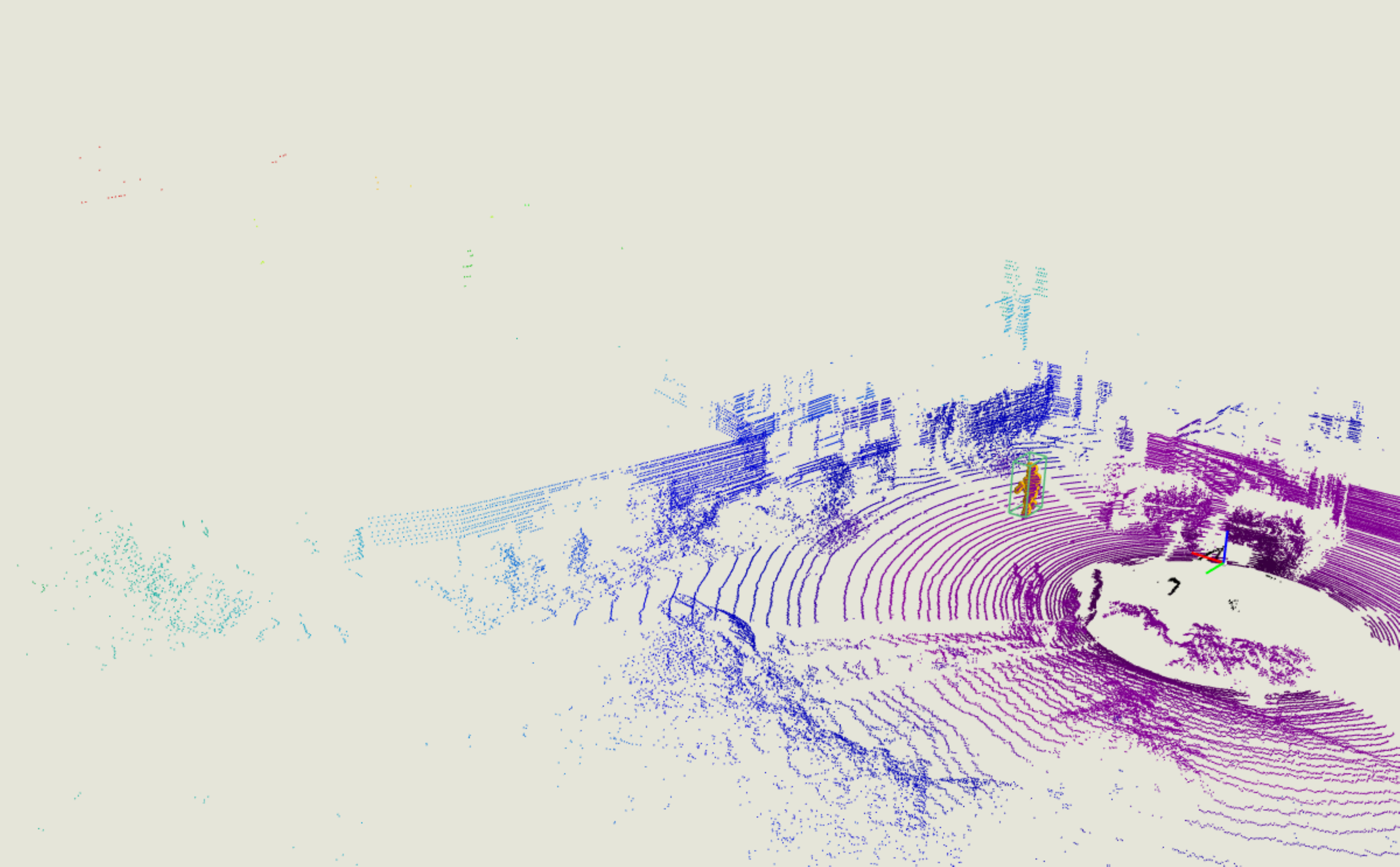}
\includegraphics[width=0.310\linewidth]{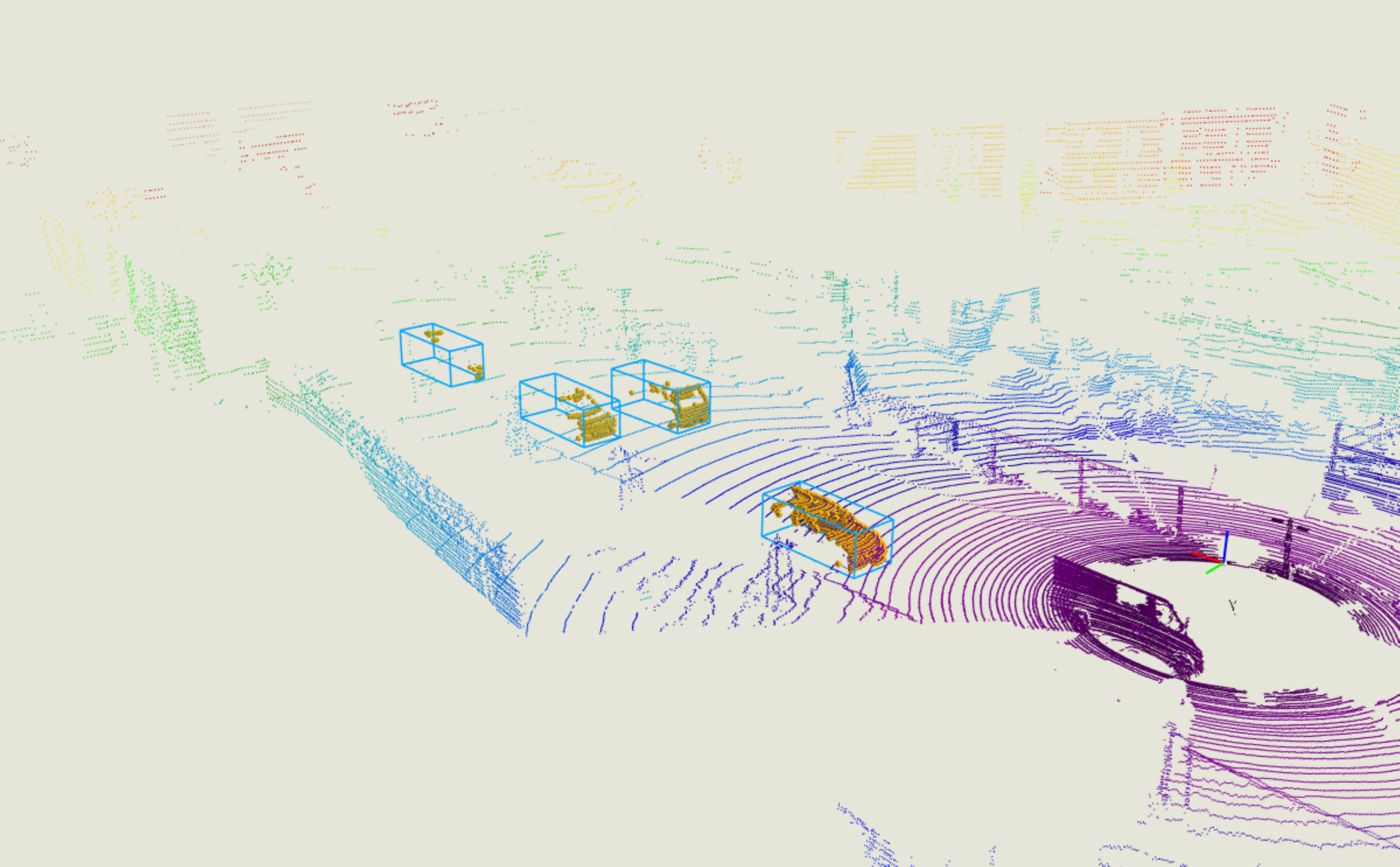}  \\ 
\vspace{1.6pt}
\includegraphics[width=0.310\linewidth]{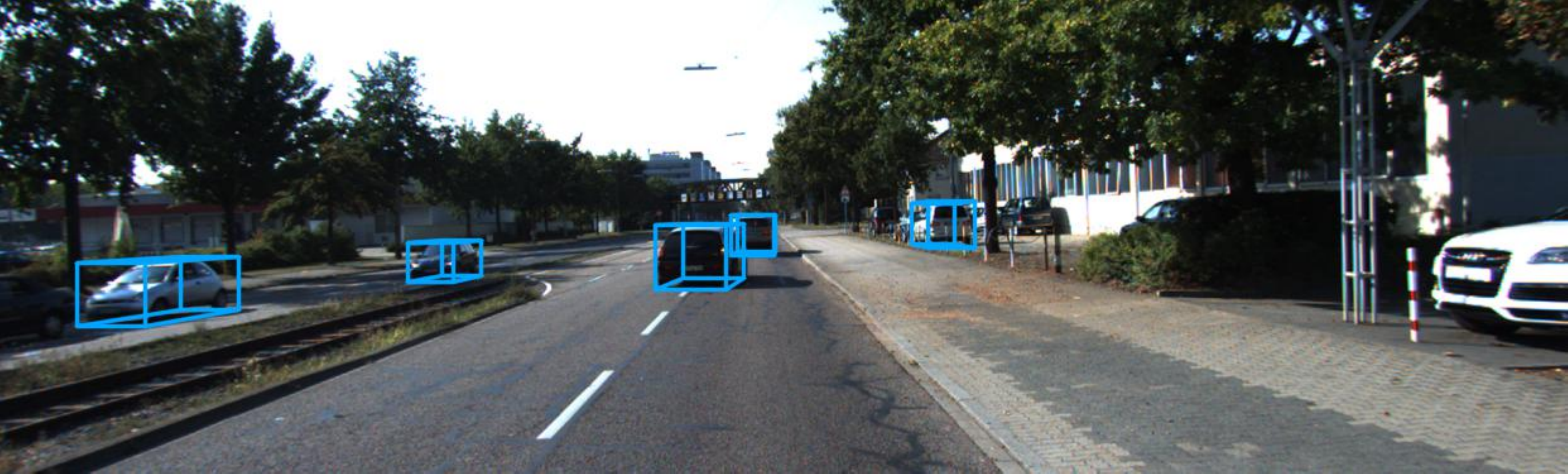}
\includegraphics[width=0.310\linewidth]{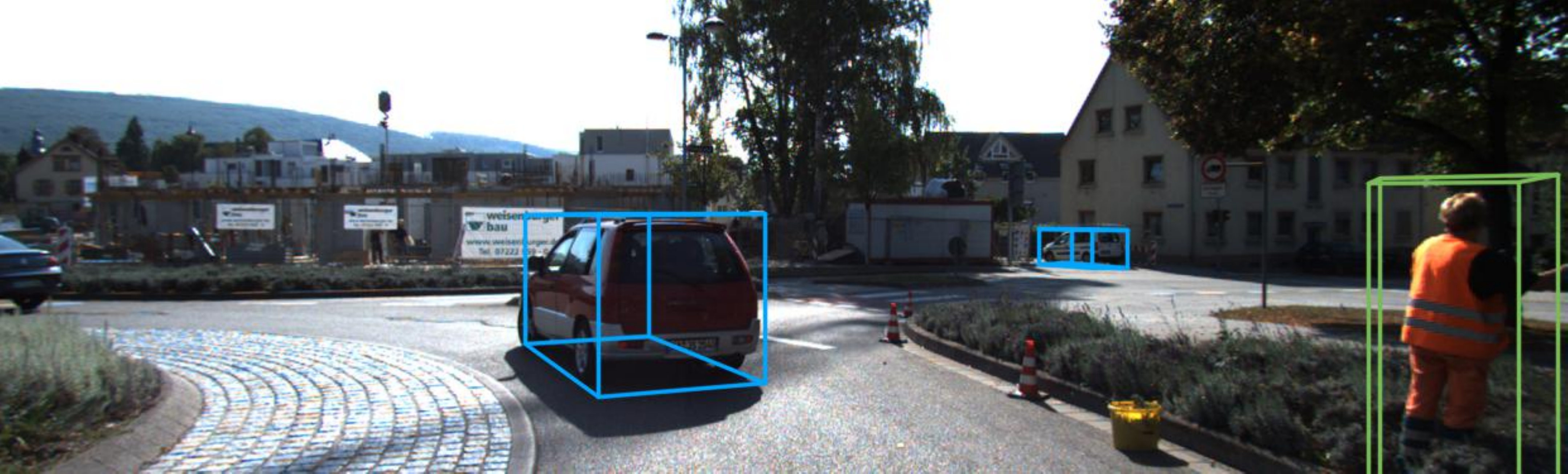}
\includegraphics[width=0.310\linewidth]{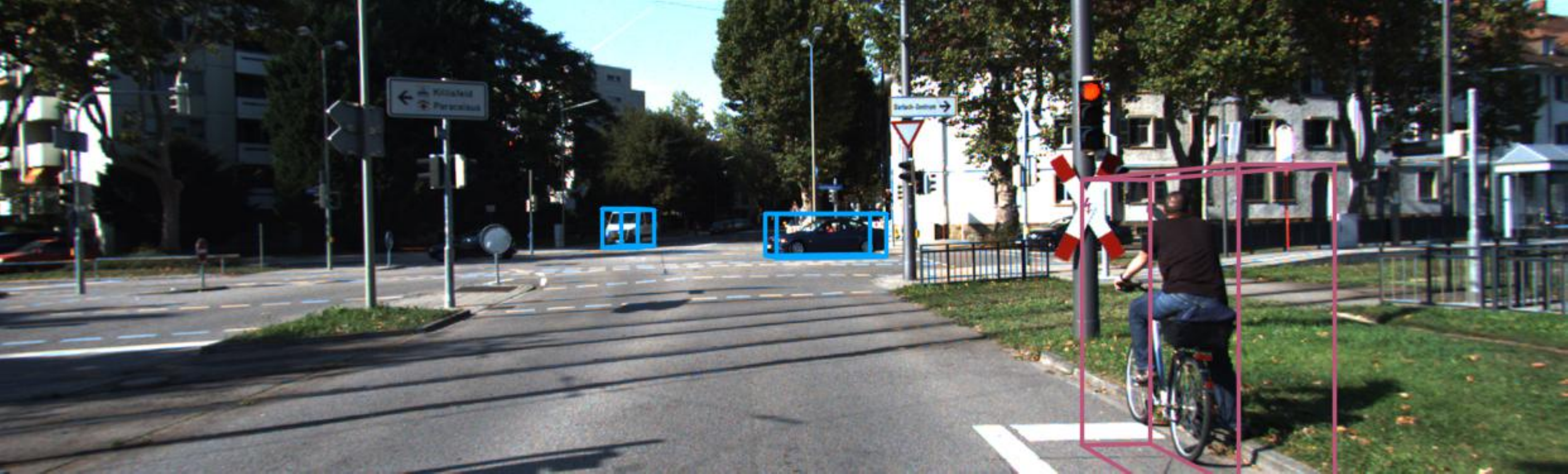} \\
\includegraphics[width=0.310\linewidth]{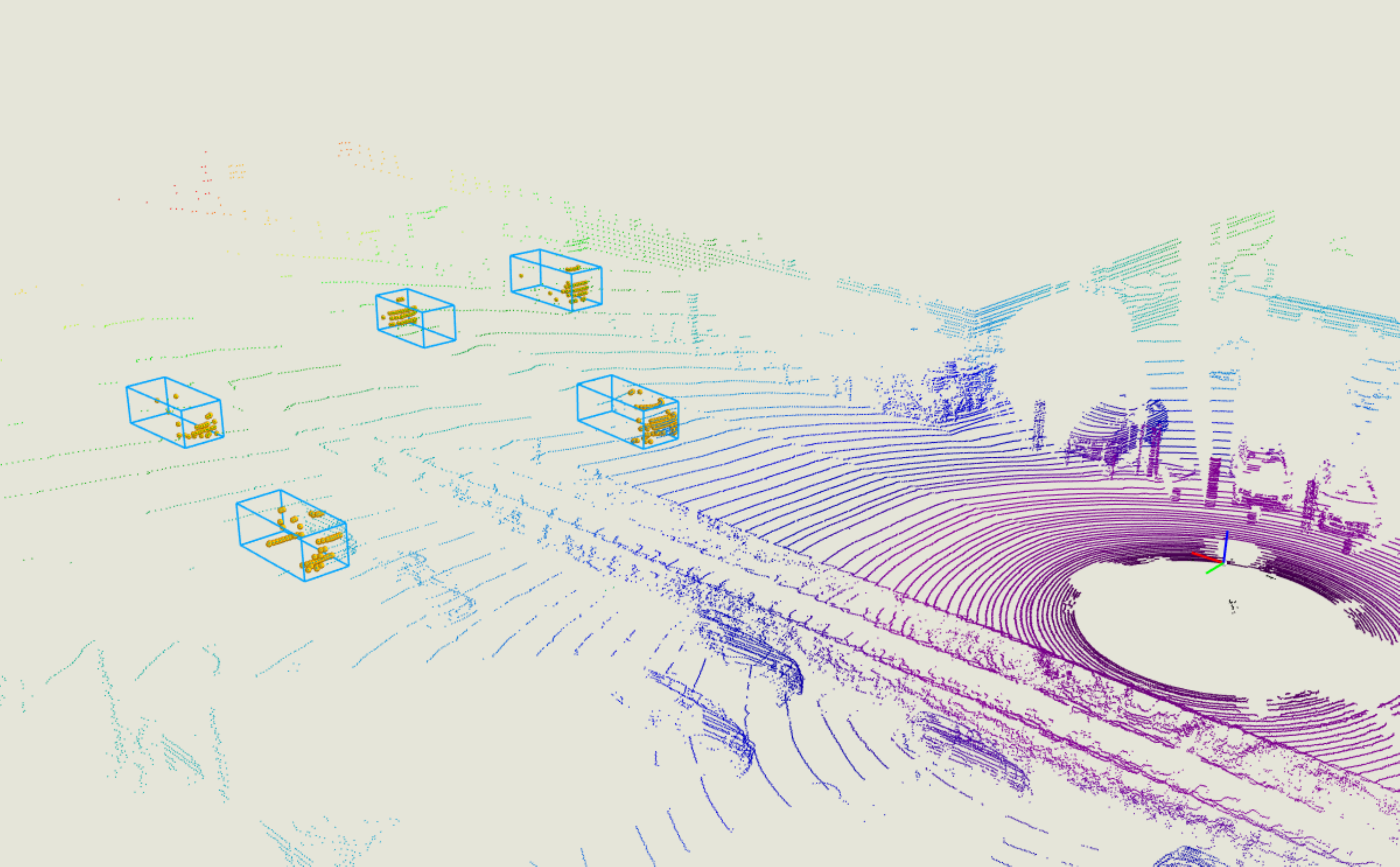}
\includegraphics[width=0.310\linewidth]{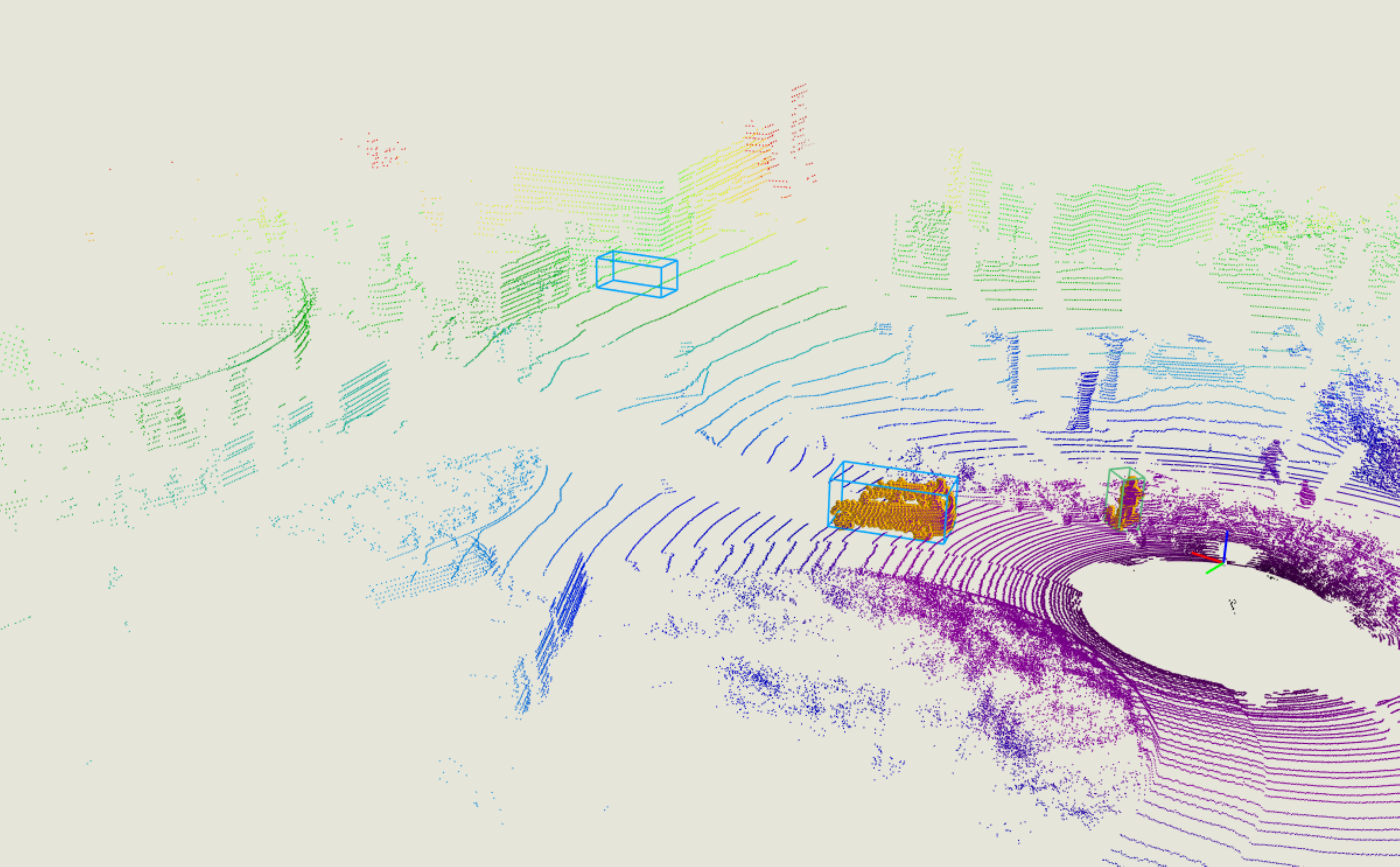}
\includegraphics[width=0.310\linewidth]{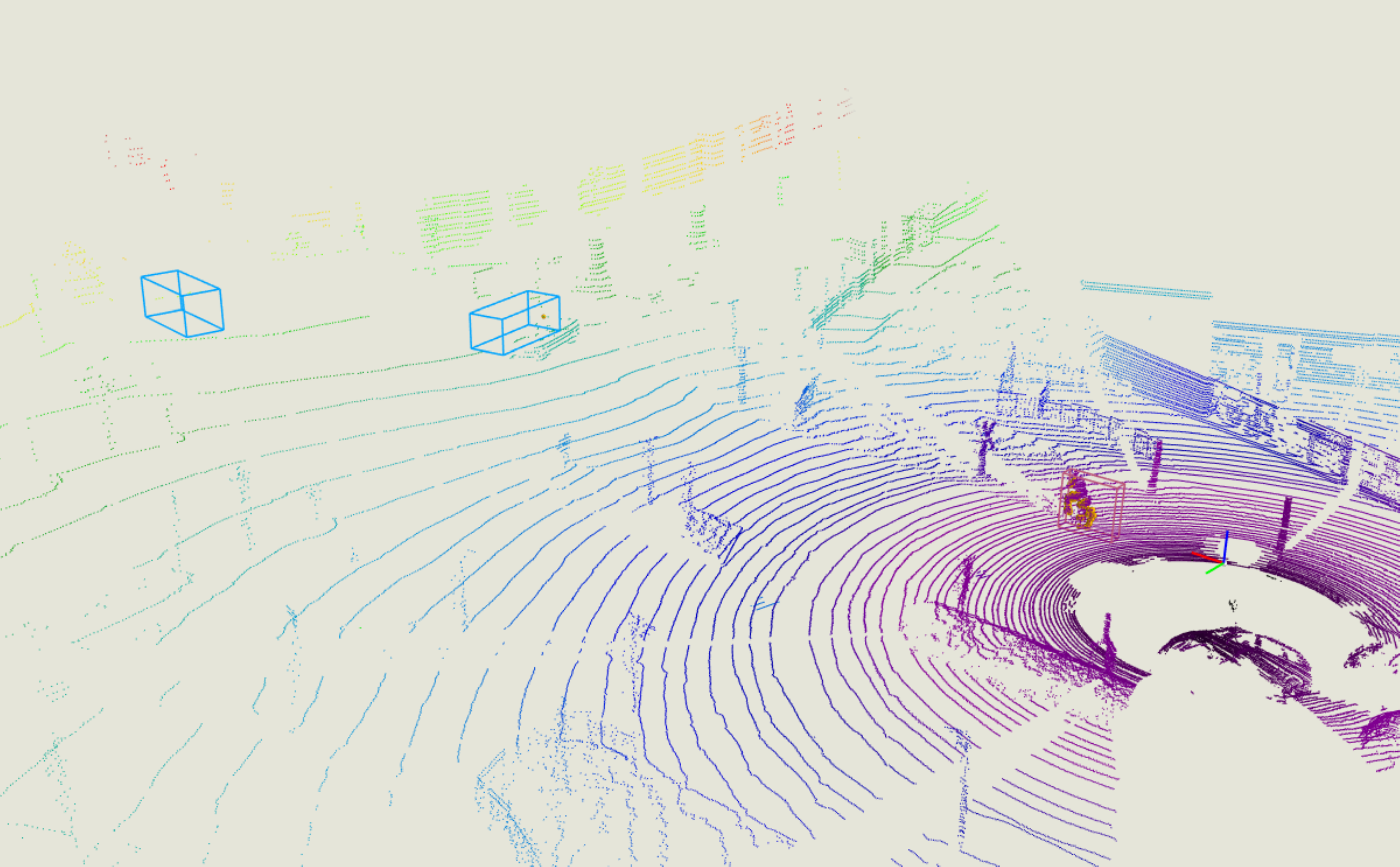}  \\
\vspace{-3pt}
   \caption{{\bf Qualitative results on the KITTI \emph{test} set.} 
   These results are based on proposed model trained on the KITTI \emph{trainval} set, running at 25 FPS.
   We use blue, green, and red boxes to denote cars, pedestrians, and cyclists. 
   LiDAR signals are only used for visualization. Best viewed in color with zoom in.}
   \label{fig:vis}
   \vspace{-8pt}
\end{figure*}

\noindent
{\bf Supervision for coarse center detection and multi-task learning.}
We show the performance changes caused by center definition and multi-task learning in Table~\ref{table:analysis_center}.
Specifically, from setting {\bf a} (used in \cite{zhou2019objects}) and setting {\bf b} (used in \cite{chen2020monopair,Simonelli_2019_ICCV}) in the table, predicting an offset to compensate for the misalignment between 2D center and projected 3D center can improve the performance of 3D detection significantly.
Then, using projected 3D center as the ground truth for coarse detection (setting {\bf d}, our model) can further improve the performance. 
Besides, by comparing the setting {\bf c} used in \cite{Liu_2020_CVPR_Workshops} and the setting {\bf d} in our design, we can find the performance of 3D detection benefits from multi-task learning (performing 2D detection and 3D detection jointly).
Note that the accuracy of 2D detection under setting {\bf d} is also better than that under setting {\bf c}, which suggests generating 2D bounding boxes from 3D detection may reduce the quality of the 2D detection results.
The above conclusions are also reflected in Table~\ref{table:testset} and \ref{table:valset}. 

\noindent
{\bf Training samples.}
From Table~\ref{table:analysis_samples}, we can find that both removing some samples from training set appropriately and reducing the training weights of them can improve overall performance.
Note that those samples are only a small part of the whole training set and will not affect the representation learning of the network to the whole dataset.
For example, in the 7,481 images in \emph{trainval} set, only 1,301/767 samples beyond 60/65 meters, accounting for 4.5\%/2.7\% of the total 28,742 samples.

\begin{table}[!t]
\centering
\begin{tabular}{l||ccc}
\toprule
   & Easy & Mod. & Hard  \\ 
\hline
baseline
& 16.12 & 12.97 & 10.99  \\
+ hard encoding, $s$ = 40 
& 14.25 & 11.25 & 9.63  \\
+ hard encoding, $s$ = 60
& 17.45 & 13.66 & 11.68 \\
+ soft encoding, $c$ = 40, $T$=1 
& 14.50 & 11.74 & 9.95 \\
+ soft encoding, $c$ = 60, $T$=1  
& 17.50 & 13.54 & 11.32 \\
+ soft encoding, $c$ = 60, $T$=5  
& 17.25 & 13.03 & 11.01 \\
\bottomrule
\end{tabular}
\vspace{-3pt}
\caption{{\bf Analysis for training samples.}
Metrics is ${\rm AP}_{40}$  of the Car category for 3D detection.}
\label{table:analysis_samples}
\vspace{-10pt}
\end{table}

\subsection{Qualitative Results}
We visualize some representative outputs of the proposed method in Figure~\ref{fig:vis}. 
To clearly show the object's position in the 3D world space, we also visualize the LiDAR signals.
We can observe that our model outputs remarkably accurate 3D bounding boxes for the cases at a reasonable distance.
We also find that our model outputs some false positive samples, \eg the 3D box on the right in the sixth picture, and the foremost reason for that is the imprecise depth or center estimation.
Note that the dimension and orientation estimation for those cases are still accurate.

%%%%   conclusion
\section{Conclusion}
In this paper, we systematically analyze the problems in monocular 3D detection and find the localization error is the bottleneck of this task.
To alleviate this problem, we first revisit the misalignment between the center of the 2D bounding box and the projected center of 3D object. 
We argue that directly detecting projected 3D center can reduce the localization error and 2D detection is conducive to optimize 3D detection.
Besides, we also find distant samples are almost impossible to detect accurately with the existing technologies, and discarding these samples from the training set will stop them from distracting the network.
Finally, we also proposed an IoU oriented loss for 3D size estimation.
Extensive experiments on the challenging KITTI dataset show the effectiveness of the proposed strategies.

\section{Acknowledgement}
This work was supported by SenseTime, the Australian Research Council Grant  DP200103223, and Australian Medical Research Future Fund MRFAI000085.

\newpage

{\small
\bibliographystyle{ieee_fullname}
\bibliography{cvpr}
}

\newpage
\setcounter{section}{0}
\renewcommand\thesection{\Alph{section}} 
\newcommand{\ym}[1]{\textcolor{blue}{#1}}
{\centering\section*{\Large Supplementary Material}}
\vspace{20 pt}

\section{Overview}
This document provides additional technical details, experimental results, theoretical analysis, and qualitative results to the main paper.
Specifically, in Section~\ref{sec:B}, we provide more details on the implementation of the depth estimation sub-task, and
Section~\ref{sec:C} shows the details and ablations about the proposed IoU oriented loss.
Section~\ref{sec:supp_discussions} provides more discussion which is omitted in the main paper.
%Experiments on a larger dataset nuScenes are presented in Section~\ref{sec:supp_nuscenes}.
Finally, Section~\ref{sec:E} presents more visual results.

\section{Depth Estimation}
\label{sec:B}
\noindent
{\bf Uncertainty modeling.}
Following~\cite{kendall2017uncertainties,kendall2019geometry,chen2020monopair}, we model the heteroscedastic aleatoric uncertainty in the depth estimation sub-task.
Specifically, we simultaneously predict the depth $\mathbf{d}$ and the standard deviation $\mathbf{\sigma}$  (or variance $\mathbf{\sigma}^{2}$):
\begin{equation}
    [\mathbf{d}, \mathbf{\sigma}] = f^{\mathbf{w}}(\mathbf{x}),
\end{equation}
where $\mathbf{x}$ is the input data and $f$ is a convolutional neural network parametrised by the parameters $\mathbf{w}$.
Then, we fix a Laplace likelihood to model the uncertainty, and the loss for the depth estimation sub-task can be formulated by: 
\begin{equation}
\mathcal{L} = \frac{\sqrt{2}}{\mathbf{\sigma}}||\mathbf{d} - \mathbf{d}^{*}||_{1} + \log\mathbf{\sigma},
\label{eq:laplacian}
\end{equation}
where $|| \cdot ||_{1}$ denotes the L1 norm and $\mathbf{d}^{*}$ is the ground truth value for depth $\mathbf{d}$.
Similarly for the Gaussian likelihood:
\begin{equation}
\mathcal{L} = \frac{1}{2\mathbf{\sigma}^{2}}||\mathbf{d} - \mathbf{d}^{*}||_{2} + \frac{1}{2}\log\mathbf{\sigma}^{2},
\label{eq:gaussian}
\end{equation}
where $|| \cdot ||_{2}$ denotes the L2 norm  (please refer to \cite{kendall2019geometry} for the derivation of Equation~\ref{eq:laplacian} and Equation~\ref{eq:gaussian}).
Note that the uncertainty modeling is not claimed as our contribution.

\begin{figure}[t]
\begin{center}
\includegraphics[width=0.99\linewidth]{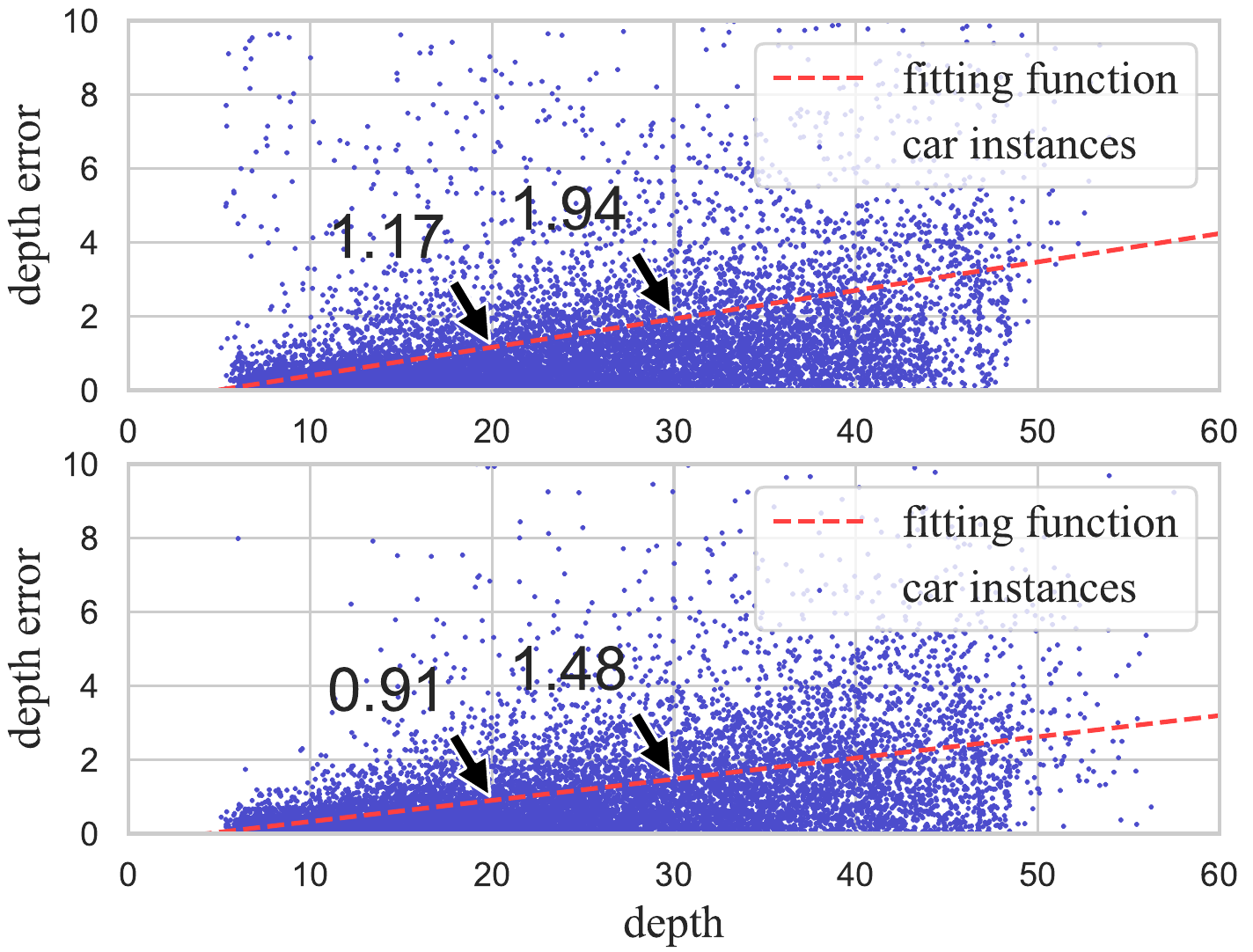}
\end{center}
\vspace{-10pt}
\caption{{\bf Errors of depth estimation.}
We show the errors of depth estimation as a function of the depth (x-axis) for the plain scheme ({\emph{top}}) and the uncertainty aware scheme based on the Laplace likelihood (\emph{bottom}).}
\label{fig:depth_uncert}
\end{figure}

\noindent
{\bf Experimental results.}
First, from Figure~\ref{fig:depth_uncert} and Table~\ref{table:depth_uncert_1}, we can find that uncertainty-based estimation improves the accuracy of depth map, thereby improving the overall performance of monocular 3D detection.
Second, the experimental result also show that modeling uncertainty based on the Laplace distribution (all models in the main paper adopted this setting) is more suitable for our task than Gaussian distribution.

\begin{table}[!t]
\centering
\resizebox{\linewidth}{!}{
\begin{tabular}{c|ccc}
\toprule
uncert. & Easy & Mod. & Hard\\ 
\hline
- & 18.56 / 13.18 & 14.24 / 10.15 & 12.13 / 8.45  \\
Gaussian & 18.68 / 13.20 & 14.22 / 10.41 & 12.08 / 8.69 \\
Laplace & 20.29 / 14.51 & 16.15 / 11.12 & 14.07 / 9.97  \\
\bottomrule
\end{tabular}}
\caption{{\bf Analysis for the designs of depth estimation.} 
Metrics are ${\rm AP}_{40}$ of the Car category for BEV/3D detection tasks.}
\label{table:depth_uncert_1}
\end{table}

\section{IoU Oriented Loss}
\label{sec:C}
\subsection{Proof of Proposition}
\noindent
This section provides the proof of the following proposition, which is used in Equations~\ref{eq:iou} and~\ref{eq:lsize} for IoU oriented optimization in Section~\ref{sec:IoU}.

\noindent
{\bf Proposition.} 
Suppose all predicted items except the 3D sizes $(h, w, l)$ are completely correct, 
the contribution ratio of each predicted side to the 3D IoU $\frac{\partial IoU}{\partial h} : \frac{\partial IoU}{\partial w} : \frac{\partial IoU}{\partial l}$ can be approximated to
$\frac{1}{h} : \frac{1}{w} : \frac{1}{l}$.

\noindent
{\it Proof.}
Given the above conditions, the 3D IoU metric can be formulated as:
\begin{equation}
\begin{small}
    IoU = \frac{\prod_{i\in\{h, w, l\}}\min(i, i^{*})}{h \times w \times l + h^{*} \times w^{*} \times l^{*} - \prod_{i\in\{h, w, l\}}\min(i, i^{*})},
\label{eq:iou_1}
\end{small}
\end{equation}
where $(h^{*}, w^{*}, l^{*})$ denotes the ground truth of 3D size $(h, w, l)$.
With the different relationship between the prediction and the ground truth of the 3D size, we can obtain the following cases:

\noindent
{\it Case 1:} If $h\leq h^{*}$, $w\leq w^{*}$, and $l\leq l^{*}$, the Equation~\ref{eq:iou_1} can be simplified as:
\begin{equation}
IoU = \frac{h \times w \times l}{h^{*} \times w^{*} \times l^{*}},
\end{equation}
and we further compute the partial derivative of 3D IoU with respect to the variable $h$ as
\begin{equation}
    \frac{\partial IoU}{\partial h} = \frac{w\times l}{h^{*} \times w^{*} \times l^{*}},
\label{eq:partial}
\end{equation}
where $ \frac{\partial IoU}{\partial h}$ represents the partial derivative of 3D $IoU$ with respect to the variable $h$, 
analogically for $\frac{\partial IoU}{\partial w}$ and $\frac{\partial IoU}{\partial l}$. Then, combining the derivative of 3D IoU with respect to $h$, $w$, and $l$, the contribution ratio of each predicted side can be given as:
\begin{equation}
\frac{\partial IoU}{\partial h} : \frac{\partial IoU}{\partial w} : \frac{\partial IoU}{\partial l}
 = \frac{1}{h} : \frac{1}{w} : \frac{1}{l}.
 \label{eq:iou_2}
\end{equation}

\noindent
{\it Case 2:} If  $h > h^{*}$, $w > w^{*}$, and $l > l^{*}$, the Equation~\ref{eq:iou_1} can be simplified as:
\begin{equation}
IoU = \frac{h^{*} \times w^{*} \times l^{*}}{h \times w \times l},
\end{equation}
and similar to Equation~\ref{eq:partial} and \ref{eq:iou_2}, we can derive the same conclusion as {\it Case 1}.

\noindent
{\it Case 3:} If  $h > h^{*}$, $w \leq w^{*}$, and $l \leq l^{*}$, then we represent the 3D IoU as:
\begin{equation}
IoU = \frac{h^{*} \times w \times l}{h \times w \times l + h^{*} \times w^{*} \times l^{*} - h^{*} \times w \times l}.
\end{equation}
By calculating the derivative of 3D IoU with respect to $h$, $w$, and $l$ respectively, we can get the contribution ratio of each predicted side:
\begin{equation}
\frac{\partial IoU}{\partial h} : \frac{\partial IoU}{\partial w} : \frac{\partial IoU}{\partial l}
 = \frac{w \times l}{h^{*} \times w^{*} \times l^{*}} : \frac{1}{w} : \frac{1}{l}.
\end{equation}

\noindent
{\it Case 4:} If  $h > h^{*}$, $w > w^{*}$, and $l \leq l^{*}$, similarly, we can get the IoU formulation as:
\begin{equation}
IoU = \frac{h^{*} \times w^{*} \times l}{h \times w \times l + h^{*} \times w^{*} \times l^{*} - h^{*} \times w ^{*}\times l}.
\end{equation}
Similar to previous steps, the formulation of each side's contribution rate to the 3D IoU is given as:
\begin{equation}
\frac{\partial IoU}{\partial h} : \frac{\partial IoU}{\partial w} : \frac{\partial IoU}{\partial l}
 = \frac{1}{h} : \frac{1}{w} : \frac{h^{*}\times w^{*} \times l^{*}}{h\times w \times l \times l}.
\end{equation}
The other cases are similar to {\it Case 3} and {\it Case 4}.
When $h\approx h^{*}$, $w\approx w^{*}$, and $l\approx l^{*}$, we can get the Equation~\ref{eq:iou} used in the main paper.

\subsection{Experiments}
We report the improvement introduced by the proposed loss function in the main paper.
To further validate the effectiveness of it, we also implement the 3D GIoU loss~\cite{3dgiou} for reference.
Specifically, we add the 3D GIoU loss as a regularization item as in ~\cite{3dgiou}, investigating different weights considered in our baseline model, and the ${\rm AP}_{40}$ of cars on the moderate setting on KITTI \emph{validation} set (Table~\ref{table:giou}) show that our IoU oriented optimization improves accuracy but 3D-GIoU with different weights does not.

\begin{table}[t]
\centering
\resizebox{\linewidth}{!}{
\begin{tabular}{c|ccc|c}
\toprule
 Baseline & GIoU (w=0.5) & GIoU (w=1) & GIoU (w=5) & Ours \\ 
\hline
  11.12 & 10.17 & 10.19 & 8.48 & 11.74\\
\bottomrule
\end{tabular}}
\caption{{\bf Ablation study} for the proposed loss function and 3D GIoU loss on the KITTI \emph{validation} set. Metric is ${\rm AP}_{40}$ of the Car category under moderate setting.}
\label{table:giou}
\end{table}

\section{Performance for the Close Objects}
\label{sec:supp_discussions}
%\subsection{Performance for the Close Objects}
The Figure~\ref{fig:evalbyrange} in the main paper provides lots of insights to us.
Except for the observations analyzed in the main paper, we also found that the performance degrades for the very close object.
Here we provides our analysis for this.
In particular, there are three main reasons in total.
a) The close-range objects tend to have larger center misalignment (see Figure~\ref{fig:stats} for the statistics). 
b) The objects at closer ranges are usually more truncated, \eg the red car (depth=3.7, truncation=0.88) and the black car (depth=6.2, truncation=0.34) in Figure~\ref{fig:feature_vis}.
c) The training samples in the close range are fewer.
For example, there are 5,979 cars in $[5m, 15m]$ and 6,707 cars in $[10m, 20m]$ on the KITTI \emph{trainval} set, and the distribution for those samples are summarized in Table~\ref{table:distribution}.
Note that the KITTI annotate the difficulty of each samples according to its size of 2D bounding box, occlusion, and truncation.
The instance with `unKnown' tag usually means that it is extremely difficult to detect and is ignored in evaluation.
With that in mind, the effective samples of those two ranges are 4,522 and 6,149.
In summary, the low performance of the very close objects is caused by the limited training samples (c) and the large proportion of hard cases (a, b).

% \section{Experiments on nuScenes}
% \label{sec:supp_nuscenes}

\begin{table}[t]
\centering
\resizebox{\linewidth}{!}{
\begin{tabular}{c|ccccc}
\toprule
Range & Easy & Moderate & Hard & UnKnown & Total\\ 
\hline
$[5m, 15m]$ & 2,131 & 1,428 & 963 & 1,457 & 5,979\\ 
$[10m, 20m]$ & 2,639 & 1,840 & 1,670 & 558 & 6,707\\ 
\bottomrule
\end{tabular}}
\caption{{\bf Data distribution} for the car samples located in $[5m, 15m]$ and $[10m, 20m]$. The data is collected from the KITTI \emph{trainval} set.}
\label{table:distribution}
\end{table}

\section{More Visualizations}
\label{sec:E}

\subsection{Learned features}
\begin{figure*}[t]
\begin{center}
\includegraphics[width=0.48\linewidth]{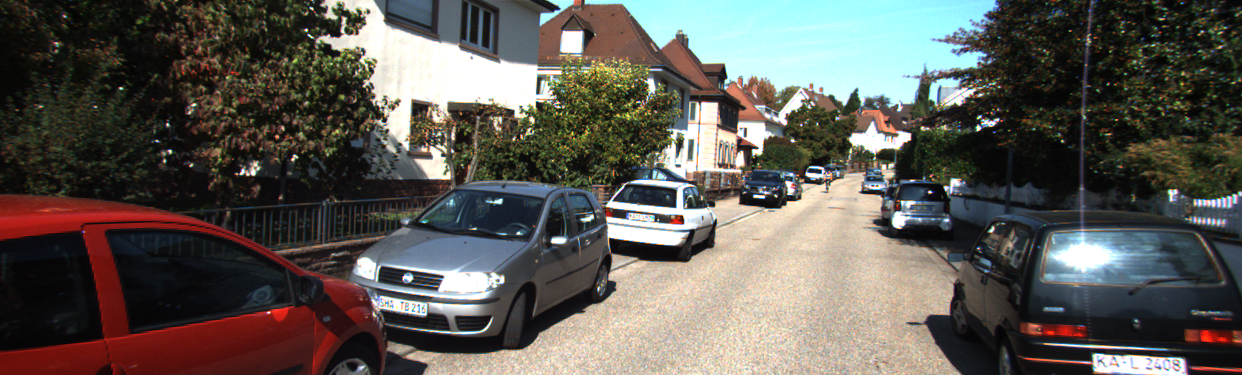}
\includegraphics[width=0.48\linewidth]{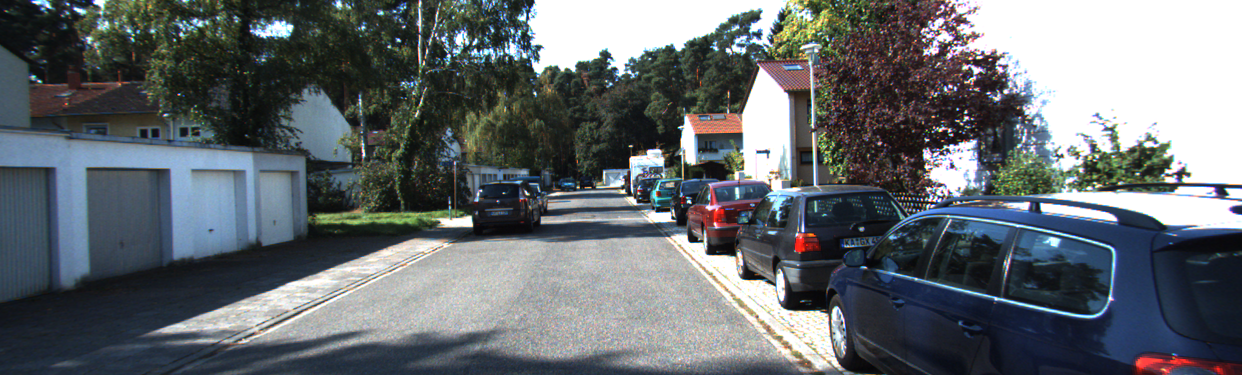}
\includegraphics[width=0.48\linewidth]{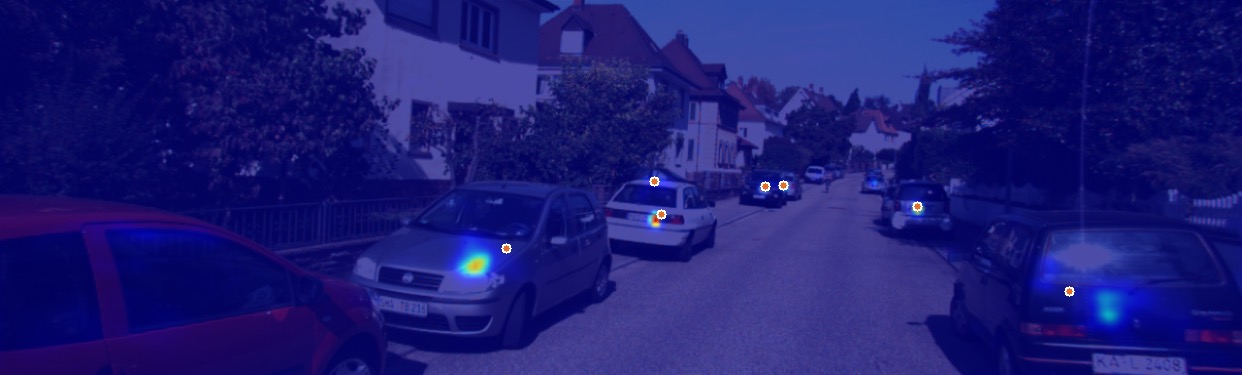}
\includegraphics[width=0.48\linewidth]{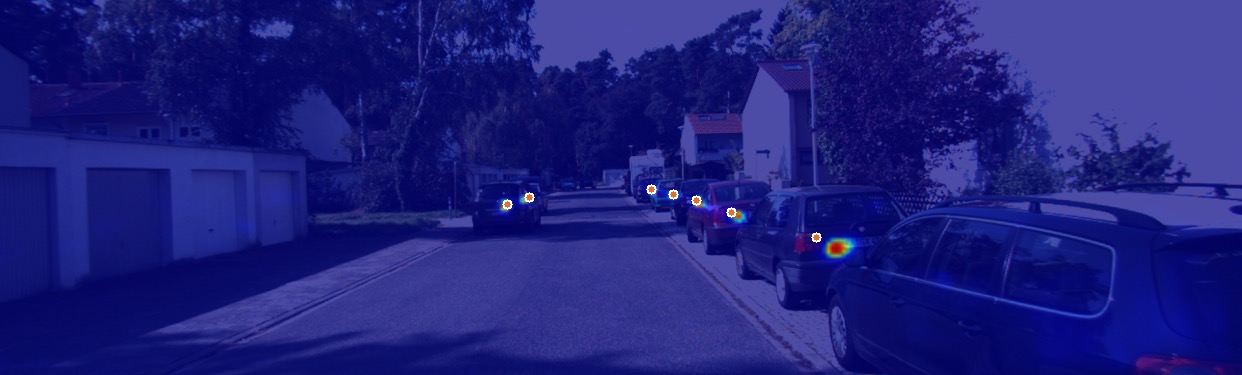}
\includegraphics[width=0.48\linewidth]{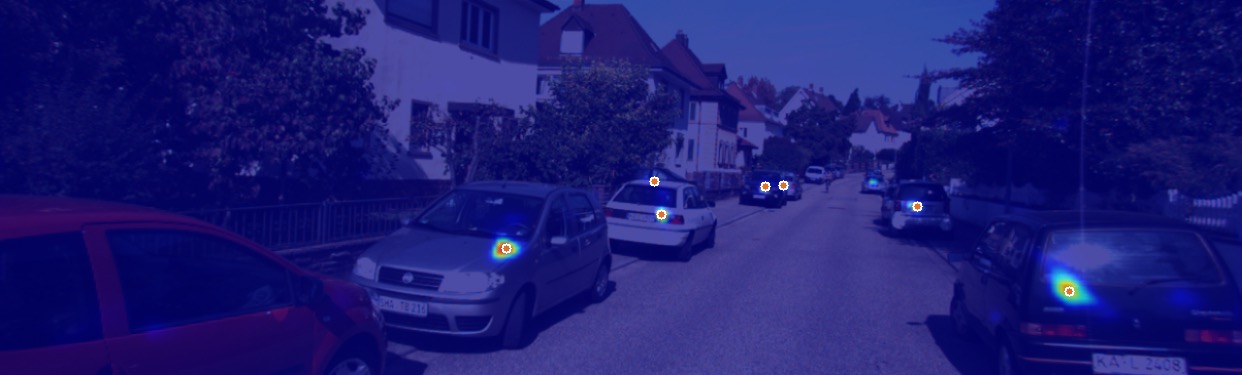}
\includegraphics[width=0.48\linewidth]{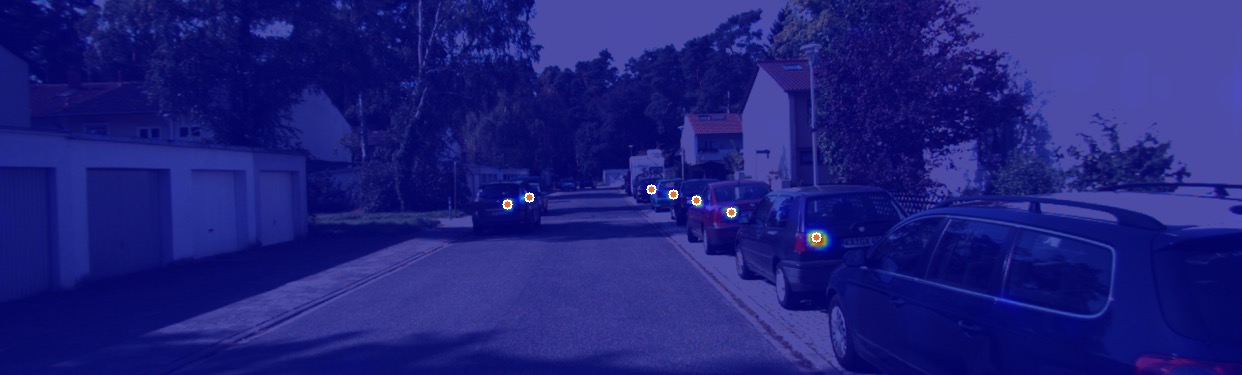}
\end{center}
\caption{{\bf Qualitative comparison for the learned features of coarse center detection task on the KITTI \emph{validation} set.} 
{\it Top:} the input image.
{\it Middle:} the features of the coarse center detection branch supervised by 2D center.
{\it Bottom:} the features of the coarse center detection branch supervised by projected 3D center.
We use the write circle to highlight the ground truth projected 3D center for better comparison.
Best viewed in color with zooming in.
}
\label{fig:feature_vis}
\end{figure*}

From Figure~\ref{fig:misalignment} in the main paper, we can see there is a misalignment between the center of the 2D bounding box and the projected center of the 3D object, especially for close objects (see Figure~\ref{fig:stats} and Figure~\ref{fig:misalignment}).
Accordingly, we propose our solution for this problem.
Here we visualize the learned features of coarse center detection branch in Figure~\ref{fig:feature_vis} to show the effectiveness of the proposed method.
The qualitative results clearly show that using projected 3D center as ground truth can make the coarse center more accurate, thereby improving the localization accuracy.

\subsection{Comparison of qualitative results}
\begin{figure*}[t]
\begin{center}
\includegraphics[width=0.48\linewidth]{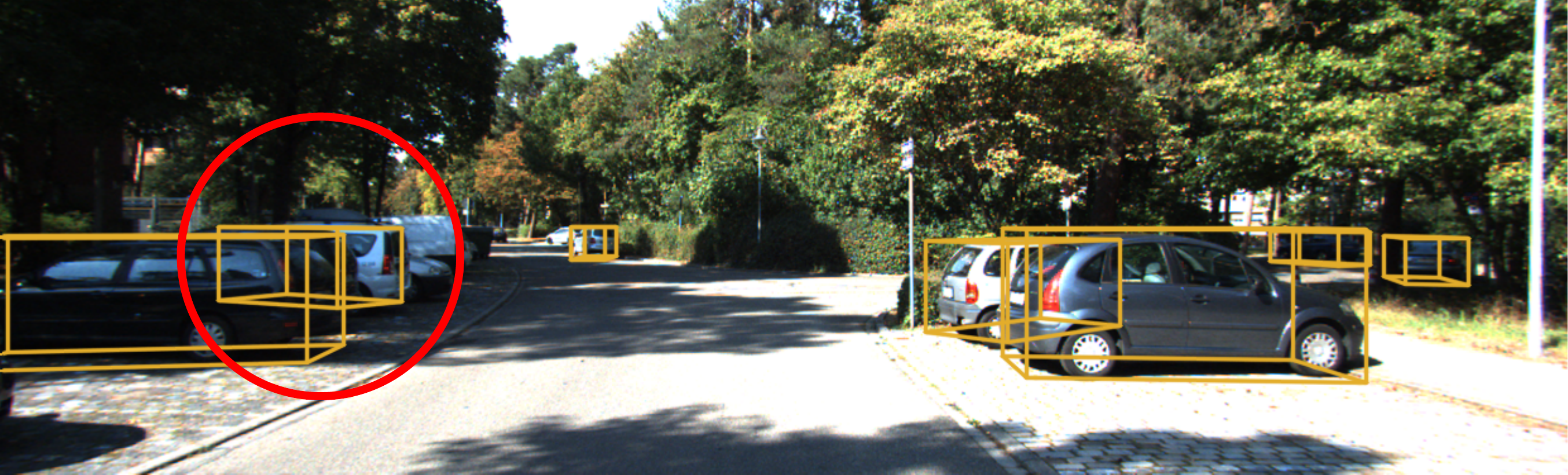}
\includegraphics[width=0.48\linewidth]{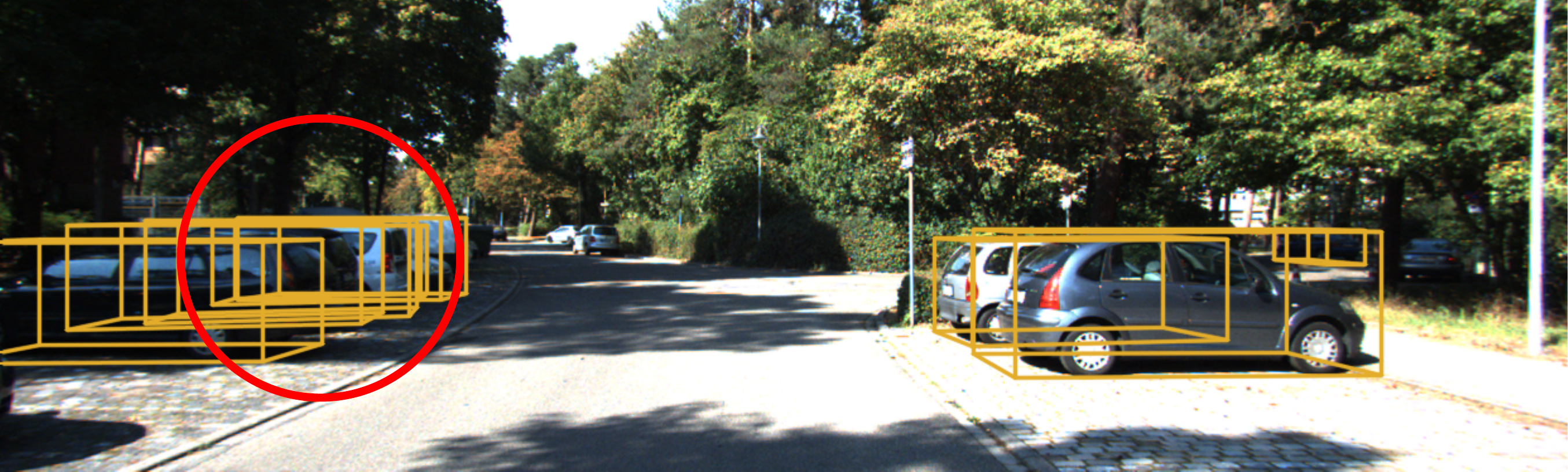}
\includegraphics[width=0.48\linewidth]{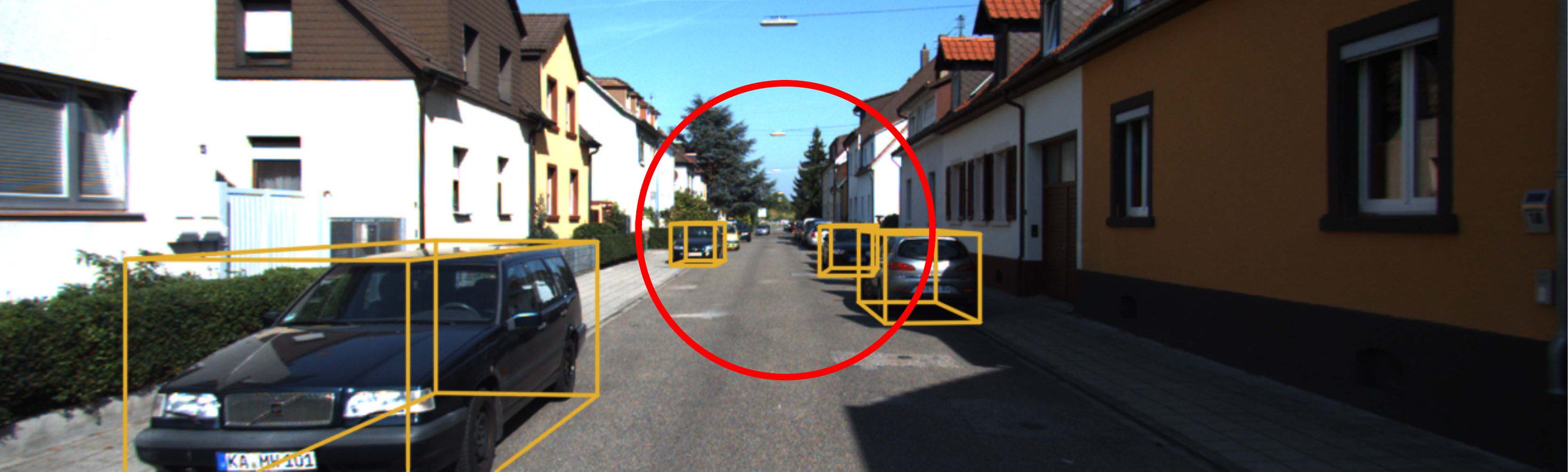}
\includegraphics[width=0.48\linewidth]{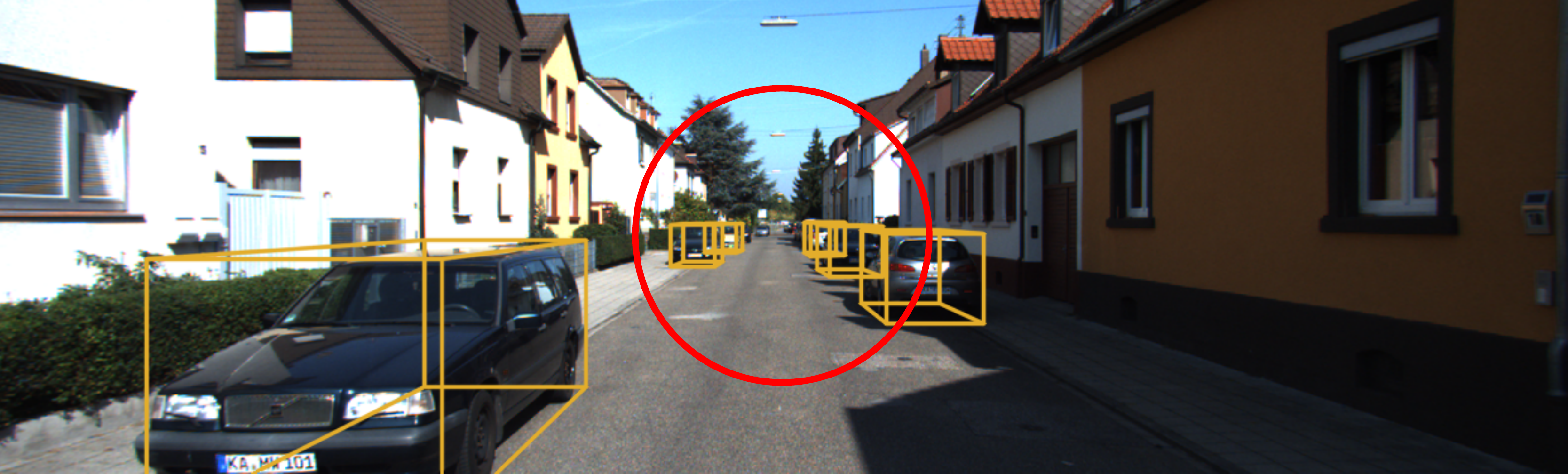}
\includegraphics[width=0.48\linewidth]{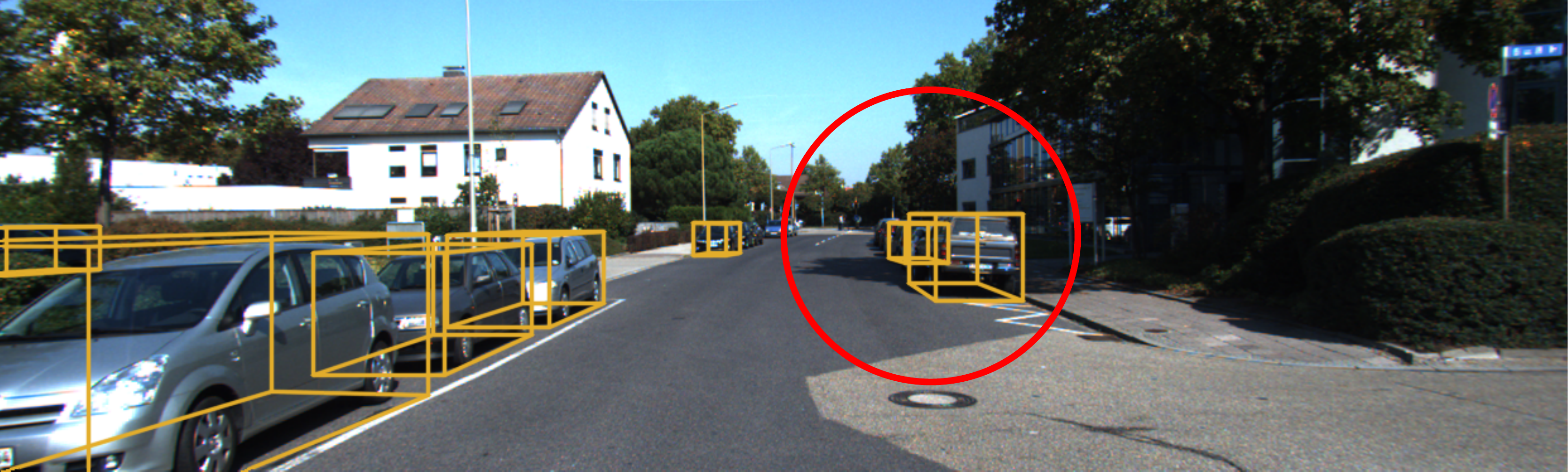}
\includegraphics[width=0.48\linewidth]{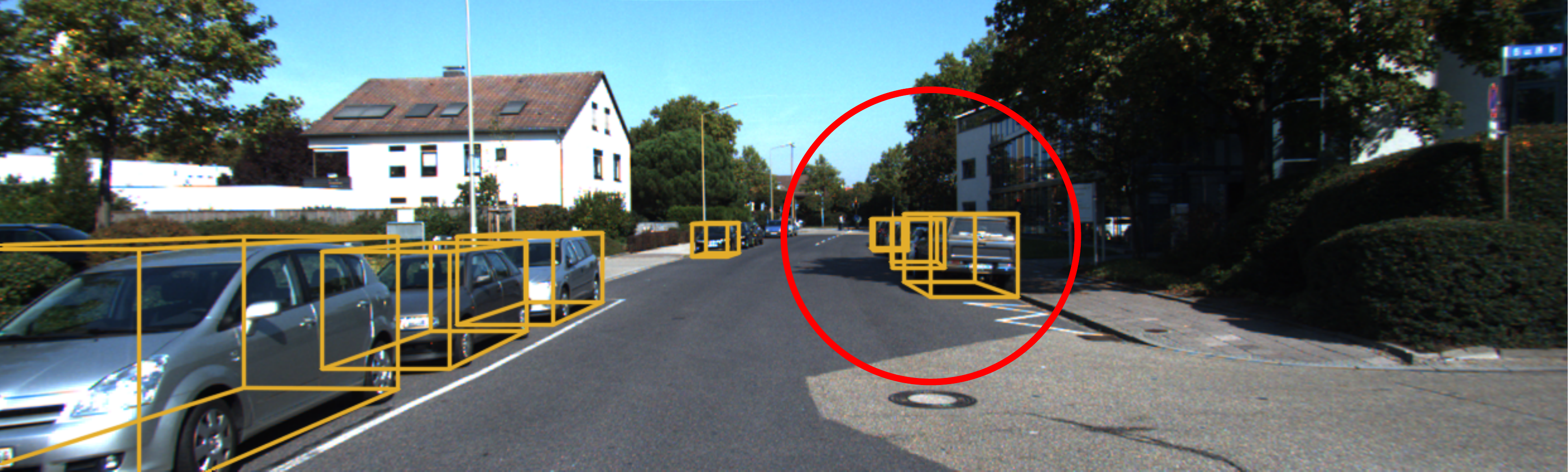}
\end{center}
\caption{{\bf Qualitative comparison on the KITTI \emph{validation} set}.
We visualize the 3D bounding boxes in the image plane.
Results are from M3D-RPN ({\it left}) and our method ({\it right}).
}
\label{fig:vis_m3d_ours}
\end{figure*}

\noindent
{\bf Visualizations in the image plane.}
We show more qualitative results of M3D-RPN (the best of all open-source standard monocular 3D detector) and the proposed method in Figure~\ref{fig:vis_m3d_ours}.
We use red circle to highlight the main differences of each pair of images, and we can find that our method performs better than M3D-RPN for dense objects.

\begin{figure*}[t]
\begin{center}
\includegraphics[width=0.48\linewidth]{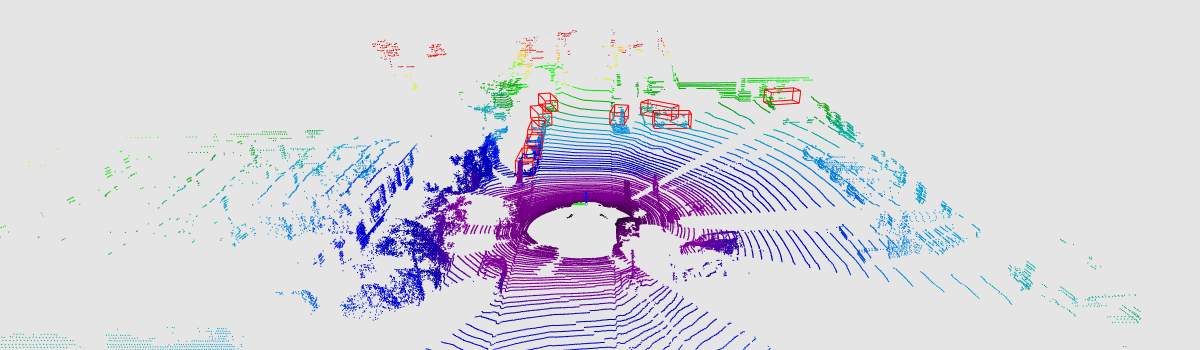}
\includegraphics[width=0.48\linewidth]{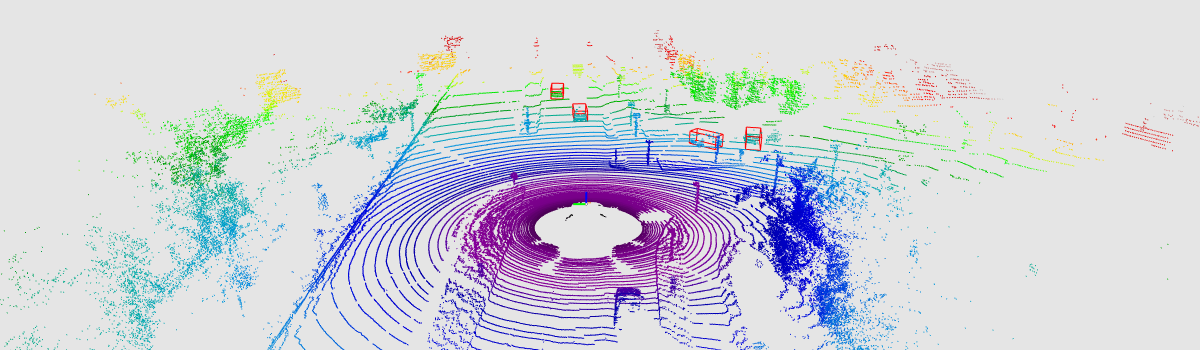}
\includegraphics[width=0.48\linewidth]{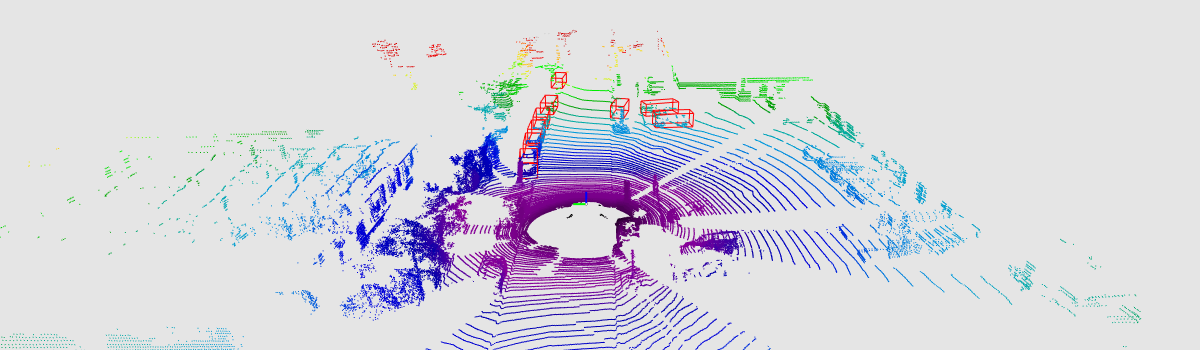}
\includegraphics[width=0.48\linewidth]{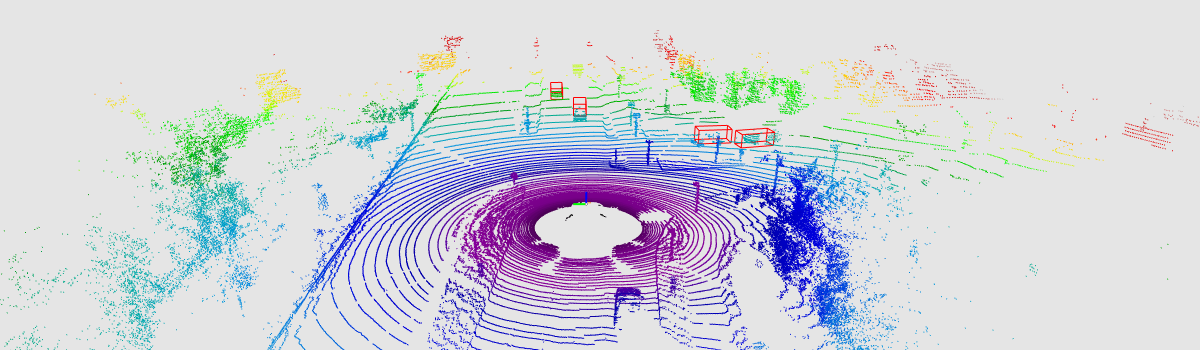}
\includegraphics[width=0.48\linewidth]{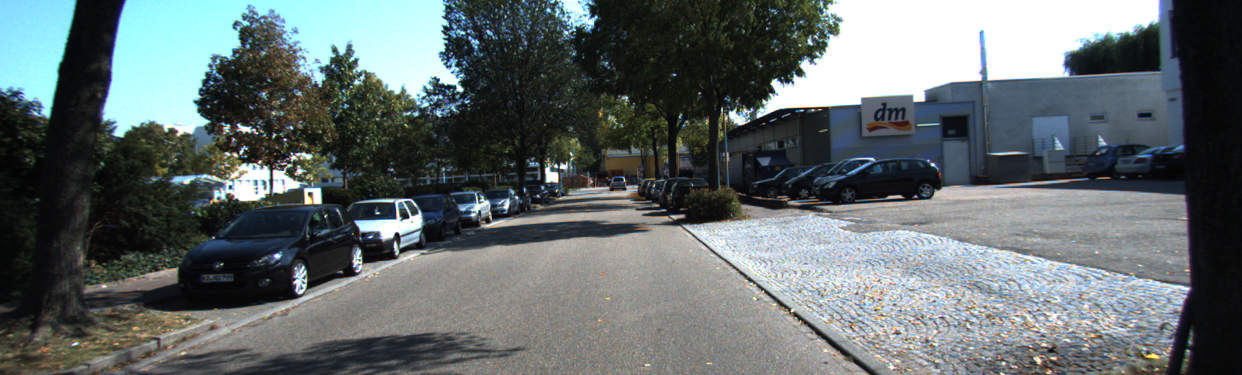}
\includegraphics[width=0.48\linewidth]{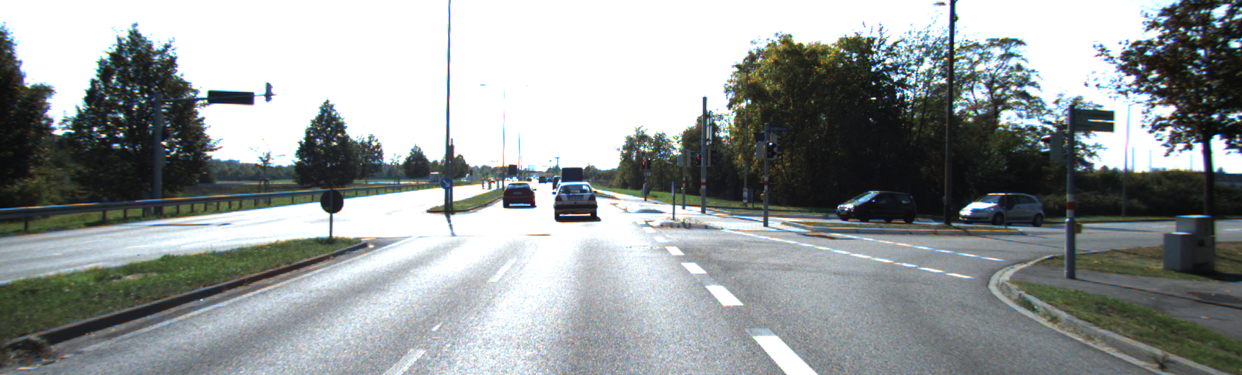}
\end{center}
\caption{{\bf Qualitative comparison on the KITTI \emph{validation} set}.
We visualize the 3D bounding boxes in the 3D world space.
Results are from M3D-RPN ({\it top}) and our method ({\it middle}).
We also show the corresponding 2D image ({\it bottom}) for reference. 
Best viewed in color with zooming in.
}
\label{fig:vis_3d_space}
\end{figure*}

\noindent
{\bf Visualizations in the 3D world space.}
We also visualize the 3D bounding boxes in the 3D world space for better presentation.
As shown in Figure~\ref{fig:vis_3d_space}, the proposed model outputs better results than M3D-RPN, especially for the orientation estimation.

\begin{figure*}[t]
\begin{center}
\includegraphics[width=0.9\linewidth]{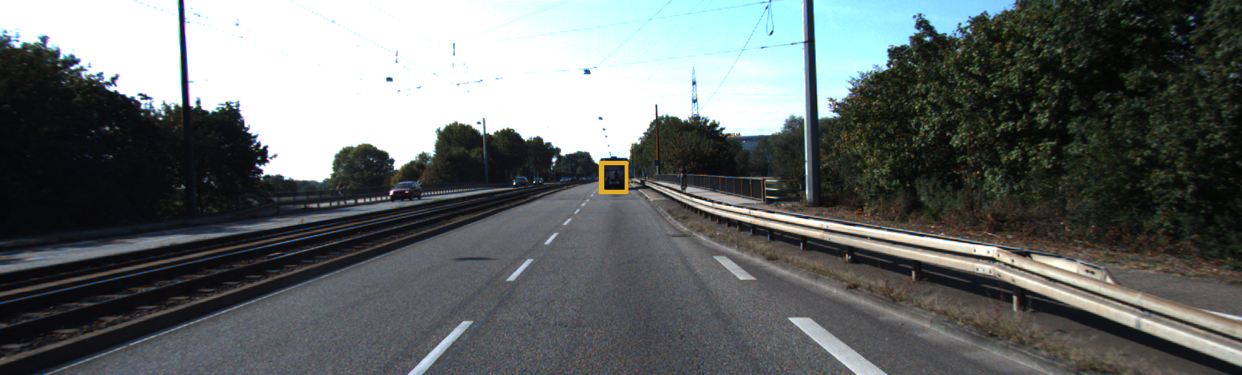}
\includegraphics[width=0.9\linewidth]{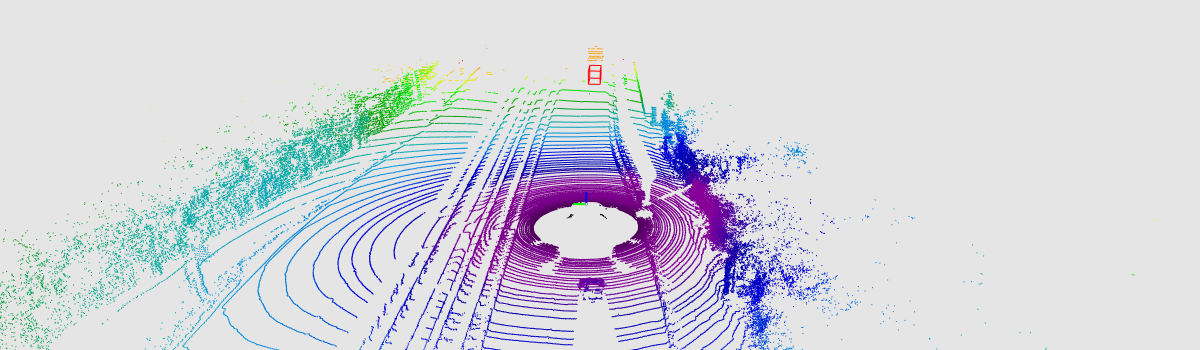}
\end{center}
\caption{{\bf Failure case}. 
We show a representative failure case which is caused by the inaccurate depth estimation.}
\label{fig:vis_error}
\end{figure*}

\noindent
{\bf Representative failure case.}
We show a typical error pattern in monocular 3D object detection in Figure~\ref{fig:vis_error}.
We can observe that the projected 3D bounding boxes fit the object's appearance tightly in the image plane. 
However, from the visualization results in the 3D world space, this is a clear false positive because the depth is inaccurate (the outline of the object can be perceived through the point clouds, best viewed with zooming in).
Note that this problem is common in the monocular 3D detection task, which  suggests that depth estimation is a key factor restricting this task.

\end{document}